%% file: main.tex
\documentclass{llncs}
\usepackage[T1]{fontenc}

\input{macros_packages}
\begin{document}
%
%\title{GenQPM: Conformal Dynamic-aware Quantitative Predictive Monitoring for Multi-Modal Scenarios}
\title{Conformal Predictive Monitoring\\ for Multi-Modal Scenarios}
\author{Francesca Cairoli\inst{1}\orcidlink{0000-0002-6994-6553} \and
Luca Bortolussi \inst{1}\orcidlink{0000-0001-8874-4001}\and Jyotirmoy V. Deshmukh\inst{2}\orcidlink{0000-0001-8874-4001} \and Lars Lindemann\inst{2}\orcidlink{0000-0001-8874-4001} \and 
Nicola Paoletti\inst{3}\orcidlink{0000-0002-4723-5363}
}

\institute{University of Trieste, Trieste, Italy \and  University of Southern California, Los Angeles, California \and King's College London, London, United Kingdom }

\maketitle    

%\linenumbers

\input{sections/abstract_new}

\input{sections/intro_new}

\input{sections/problem}

\input{sections/background}

\input{sections/methods}

\input{sections/experiments}

\input{sections/related}

\input{sections/conclusions}

\vspace{3mm}

\noindent \textbf{Acknowledgments.} 
This work has been partially supported by the ``REXASI-PRO'' H-EU project, call HORIZON-CL4-2021-HUMAN-01-01, Grant agreement ID: 101070028 and by the PNRR project iNEST (Interconnected North-Est Innovation Ecosystem) funded by the European Union Next-GenerationEU (Piano Nazionale di Ripresa e Resilienza (PNRR) – Missione 4 Componente 2, Investimento 1.5 – D.D. 1058 23/06/2022, ECS\_00000043).

\bibliographystyle{splncs04}
\bibliography{references}

%\clearpage
%\input{sections/appendix}

\end{document}

%% file: macros_packages.tex
\usepackage{hyperref}
\usepackage{graphicx} 
\usepackage{mathtools}
\usepackage{mathrsfs}
\usepackage{physics}
\usepackage{xcolor}
\usepackage{url}

\usepackage{amssymb}

\usepackage{algorithm}

\usepackage{algpseudocode}
\usepackage{lineno}
\usepackage{orcidlink}

\usepackage{wrapfig}

\usepackage{adjustbox}
\usepackage{array}
\usepackage{booktabs} % Per linee di tabella migliorate

\usepackage{tikz}
\usepackage[toc,page]{appendix}

\newcommand{\camera}[1]{#1}

\newcommand{\blind}[1]{}
\newcommand{\new}[1]{#1}

\newcommand{\rv}[1]{#1}

\newcommand{\horizon}{H}

\newcommand{\inp}{\mathbf{x}}
\newcommand{\targ}{t}
\newcommand{\init}{\vec{s}(0)}%c
\newcommand{\initi}{\vec{s}_i(0)}
\newcommand{\latent}{z}
\newcommand{\traj}{\vec{s}}% \mathbf{y}
\newcommand{\statespace}{S}

\newcommand{\condist}{p(\traj | \init)}
\newcommand{\weights}{\theta}
\newcommand{\gendist}{p_\weights(\traj | \init)}
\newcommand{\gen}{\texttt{gen}}
\newcommand{\genfnc}{\texttt{gen}_\weights(\latent , \init)}
\newcommand{\gendisti}{p_\weights(\traj | \initi)}
\newcommand{\diffsteps}{\mathcal{T}}
\newcommand{\partition}{\mathcal{M}}
\newcommand{\predinterval}{\texttt{Int}}
\newcommand{\rob}{\texttt{Rob}_\phi}

\newcommand{\mode}{\texttt{mode}}

\newcommand{\prob}{Pr}
\newcommand{\distrib}{p}

\newcommand{\ourmethod}{GenQPM}

%% file: sections/abstract_new.tex
We consider the problem of quantitative predictive monitoring (QPM) of stochastic systems, i.e., predicting at runtime the degree of satisfaction of a desired temporal logic property from the current state of the system. Since computational efficiency is key to enable timely intervention against predicted violations, several state-of-the-art QPM approaches rely on fast machine-learning surrogates to provide prediction intervals for the satisfaction values, using conformal inference to offer statistical guarantees. 
However, these QPM methods suffer when the monitored agent exhibits multi-modal dynamics, whereby certain modes may yield high satisfaction values while others critically violate the property. Existing QPM methods are mode-agnostic and so would yield overly conservative and uninformative intervals that lack meaningful mode-specific satisfaction information. 
To address this problem, we present \ourmethod, a method that leverages deep generative models, specifically score-based diffusion models, to reliably approximate the probabilistic and multi-modal system dynamics without requiring explicit model access. \ourmethod\ employs a mode classifier to partition the predicted trajectories by dynamical mode. For each mode, we then apply conformal inference to produce statistically valid, mode-specific prediction intervals. 
We demonstrate the effectiveness of \ourmethod\ on a benchmark of agent navigation and autonomous driving tasks, resulting in prediction intervals that are significantly more informative (less conservative) than mode-agnostic baselines.

%% file: sections/intro_new.tex
\section{INTRODUCTION}\label{sec:intro}
\textit{Predictive monitoring} is an advanced form of runtime verification that leverages forecasts of a system's future behavior to detect safety violations before they occur. Traditional runtime verification checks whether a system trace satisfies a temporal logic property in real time but only considers the observed prefix of the trajectory, i.e., it cannot reason about possible future evolutions of the system~\cite{leucker2009brief}. In contrast, predictive monitoring incorporates a model of system dynamics to predict future trajectories and assess, in an online manner, whether a violation is imminent from the current state, thereby enabling preemptive and timely safety interventions.

This paper focuses on predictive monitoring for \textit{stochastic systems}, where future behavior is inherently uncertain. In such systems, the satisfaction of temporal logic specifications becomes a stochastic quantity (some evolutions can be safe, others unsafe). Although probabilistic model checking (PMC)~\cite{baier2008principles} can provide exact satisfaction probabilities, it requires full knowledge of the system model and becomes intractable for large or continuous state spaces. Statistical Model Checking (SMC)~\cite{younes2006statistical} alleviates some of these problems by estimating satisfaction probabilities from sampled trajectories, but it must be run at each state during execution. This makes both PMC and SMC impractical for online predictive monitoring. In particular, we focus on correctness specifications given as \textit{Signal Temporal Logic (STL)} formulas~\cite{maler2004monitoring,donze2010robust} and on monitoring STL robustness, i.e., the logic’s quantitative semantics indicating the degree of satisfaction of an STL formula by a trajectory. We call this problem \textit{Quantitative Predictive Monitoring (QPM)}. 
Our work builds on recent QPM methods~\cite{cairoli2023conformal,lindemann2023conformal,zhao2024robust,cairoli2025conformal}, which rely on machine-learning models for trajectory predictions and employ the conformal prediction framework~\cite{vovk2005algorithmic,angelopoulos2021gentle} to produce prediction intervals for STL robustness with guaranteed coverage. However, these methods fail to provide useful results in scenarios with \textit{multi-modal dynamics}.

\begin{figure}[!t]
    \centering
    \vspace{-.6cm}
\includegraphics[width=0.8\linewidth]{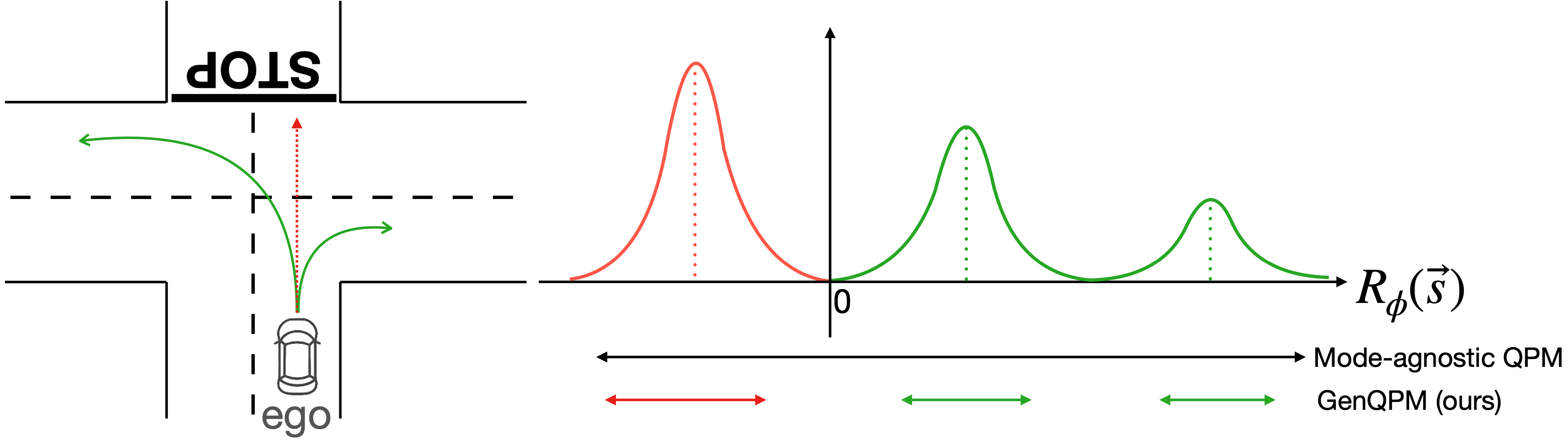}
\vspace{-.25cm}
\caption{ Left: Vehicle facing multiple possible choices. Right: Distribution of the robustness values of these choices/modalities w.r.t. an STL safety requirement. Bottom-right: uninformative mode-agnostic prediction interval vs informative mode-specific intervals.\vspace{-.5cm}}\label{fig:multimodality}

\end{figure}
The main concern in this paper is the predictive monitoring of multi-modal systems, i.e.,  systems that can evolve along qualitatively different future behaviors (or ``modes’’). For instance, consider the example in Fig.~\ref{fig:multimodality}-left, where we see an autonomous vehicle approaching an intersection. The vehicle might turn left, turn right, or go straight --- each representing a distinct mode. The first two modes lead to safe scenarios, i.e., scenarios that satisfy the safety requirements (positive STL robustness), whereas the latter mode violates safety/traffic rules (negative STL robustness). 
Fig.~\ref{fig:multimodality}-right highlights the multi-modal distribution of STL robustness values in such an environment. In such settings, the above-mentioned methods would fail to capture the impact of different modes.
So, they would produce overly conservative  regions that cover a broad range of outcomes, losing valuable information about which modes are safe or unsafe.

To address this limitation, our goal is to develop a predictive monitor that can quantify the mode-specific satisfaction of STL requirements. That is, rather than providing a single large prediction interval over possible robustness values, we aim to produce separate prediction intervals for each mode, offering insights into which modes are likely to lead to violations or satisfaction, thus enhancing the decision-making process of the agent.

To this purpose, we propose \textit{\ourmethod}, a novel  \textit{dynamics-aware quantitative predictive monitor} that leverages deep generative models and conformal inference to provide reliable and mode-specific prediction intervals. First, we use \textit{score-based diffusion models}~\cite{ho2020}, a class of state-of-the-art generative models, to learn a probabilistic surrogate of the stochastic system dynamics and generate possible future trajectories. Second, we partition these trajectories into distinct dynamical modes using a \textit{mode classifier}, which may be known or learned from data (in either a supervised or unsupervised fashion). For each mode, we estimate the distribution of STL robustness values and compute prediction intervals by applying \textit{conformalized quantile regression (CQR)}~\cite{romano2019conformalized}. Doing so, our approach ensures statistically valid coverage, even if the generative surrogate model is approximate. 
In summary, the main contributions of this paper are:
\begin{enumerate}
\item We introduce \ourmethod, a method that combines diffusion models with conformal inference to enable STL predictive monitoring for multi-modal stochastic systems.
\item \ourmethod\ provides probabilistic guarantees and supports both known and learned mode predictors.
\item We evaluate \ourmethod\ on four case studies, including a 2D navigation task and a multi-agent cross-road scenario. Our method yields mode-specific robustness intervals that are more interpretable and substantially tighter than mode-agnostic QPM baselines.
\end{enumerate}

%% file: sections/problem.tex
\section{PROBLEM STATEMENT}\label{sec:problem}

% \rv{After introducing stochastic processes and signal temporal logic, we present our predictive monitoring problem.}

%%--------------- STOCH PROCESS ----------
\paragraph{Stochastic Process.} 
\rv{We consider systems that can be described as stochastic processes (as in~\cite{cairoli2023conformal}), defined by a collection of random variables ${\mathbf{S}(i,\omega)}_{i\in\mathcal{I},\camera{\omega\in\Omega}}$ over a probability space $(\Omega,\mathcal{F},\mathbb{P})$. \camera{Here, $\Omega$ is the sample space (all possible outcomes), $\mathcal{F}$ is the event space (all subsets of $\Omega$), and $\mathbb{P}$ is the probability function, assigning each event a value in $[0,1]$}. $\mathcal{I}={0,1,\ldots}$ is a discrete-time index set, and each $\mathbf{S}(i,\omega)$ maps to a measurable state space $S\subseteq\mathbb{R}^n$. A realization $\mathbf{S}(\cdot,\omega): \mathcal{I}\rightarrow S$ is a sample trajectory. For a horizon $H$, $\Vec{s}$ denotes a realization of the random sequence $\Vec{\mathbf{S}}=(\mathbf{S}(0,\cdot), \ldots, \mathbf{S}(H-1,\cdot))$, where $\mathbf{S}(k,\cdot)$ is the state -- a random variable -- at time $k$.}

\paragraph{Signal Temporal Logic (STL).}\label{subsec:stl}
\rv{STL~\cite{maler2004monitoring,donze2010robust} specifies properties of real-valued signals via the syntax:
$\phi := \mathit{true}\ |\ \mu\ |\ \neg\phi\ |\ \phi\land\phi\ |\ \phi\ U_I\ \phi$,
where predicates $\mu \equiv g(\cdot)>0$ are evaluated based on a predicate function $g:S\to\mathbb{R}$ and where $I=[a,b]\subseteq\mathbb{T}$ are time intervals. Derived operators include $\mathit{false}$, $\lor$, $F_I\phi$ (eventually), and $G_I\phi$ (globally). The Boolean semantics is defined as follows:
\begin{align*}
&(\Vec{s}, t) \models \mu &\Leftrightarrow &\hspace{2ex} g(\Vec{s}(t))>0 \vspace{0.02in}\\
&(\Vec{s}, t) \models \neg \phi &\Leftrightarrow &\hspace{2ex} \neg((\Vec{s}, t) \models \phi)\vspace{0.02in}\\
&(\Vec{s},t) \models \phi_1 \wedge \phi_2  &\Leftrightarrow&\hspace{2ex} (\Vec{s},t) \models \phi_1 \wedge (\Vec{s},t) \models \phi_2 \vspace{0.02in}\\
&(\Vec{s},t) \models \phi_1 U_{[a,b]} \phi_2 &\Leftrightarrow&\hspace{2ex} \exists\ t'\in [t + a,t+b]\ \text{s.t.}\ (\Vec{s},t') \models \phi_2\
\forall\; t''\in [t,t'],\ (\Vec{s},t'') \models \phi_1 
\end{align*}
\camera{Let $S^H$ denote the space of trajectories $\Vec{s}$ of length $H$ and let $\mathcal{I}_H=\{0,1,\dots, H-1\}$ denote the discrete time interval, the} quantitative semantics, or robustness, is defined as a function $\texttt{Rob}: S^H\times \camera{\mathcal{I}_H} \rightarrow \mathbb{R}$:
\begin{align*}
    &  \texttt{Rob}_\mu (\Vec{s},t) = \ g(\Vec{s}(t))\\
    &  \texttt{Rob}_{\neg\phi}(\Vec{s},t)= - \texttt{Rob}_{\phi}(\Vec{s},t)\\
    &  \texttt{Rob}_{\phi_1\land\phi_2}(\Vec{s},t) = \min ( \texttt{Rob}_{\phi_1}(\Vec{s},t), \texttt{Rob}_{\phi_2}(\Vec{s},t))\\
    & \texttt{Rob}_{\phi_1 U_{[a,b]}\phi_2}(\Vec{s},t) =
    \underset{\scriptscriptstyle t'\in [t+a,t+b]}{\sup}\Big(\min\big( \texttt{Rob}_{\phi_2}(\Vec{s},t'), \underset{\scriptscriptstyle t''\in [t,t']}{\inf} \texttt{Rob}_{\phi_1}(\Vec{s},t'')\big)\Big).
\end{align*}
The robustness of a trajectory quantifies the level of satisfaction with respect to $\phi$. Positive robustness means that the property is satisfied, whereas negative robustness means that the property is violated. In this paper, we focus on the above quantitative semantics, but our framework could accommodate 
alternative semantics as well (e.g., temporal robustness~\cite{rodionova2021time} or space-time robustness~\cite{donze2010robust}).}

\subsubsection*{Quantitative Predictive Monitoring}
Given a stochastic process $\mathbf{S} = \{\mathbf{S}(t,\omega),t\in \mathcal{I}\}$, an STL requirement $\phi$ and a state $s_0\in S$, the robustness over future evolutions of the system starting from $s_0$ is distributed according to the conditional distribution $\mathbb{P}_{{\Vec{s}\sim\Vec{\mathbf{S}}}}\big(\rob(\Vec{s},0) \mid \Vec{s}(0)=s_0\big)$. 
This conditional distribution captures the distribution of the STL robustness values for trajectories of length $H$ starting in $s_0$.

 \rv{The quantitative predictive monitoring (QPM) problem can be expressed as follows. From any state $\vec{s}(0)$ of the \emph{unknown} stochastic process, we aim to construct a prediction interval guaranteed to include, with arbitrary probability, the true STL robustness of any (unknown) stochastic trajectory starting at $\vec{s}(0)$. %A formal statement of the problem is given below and 
 QPM builds on the only assumption of having access to a set $D$ of observed realizations of the random sequence $\Vec{\mathbf{S}}$, i.e., a set of trajectories of length $H$ coming from an unknown stochastic process $\mathbf{S}$. The trajectories in $D$ are split into a training set $D_t$ and a calibration set $D_c$.
 Formally, the guarantees offered by QPM in~\cite{cairoli2023conformal} can be expressed as:  for a significance level $\alpha\in (0,1)$
 \begin{equation}\label{eq:marg_cov}    \mathbb{P}_{{\Vec{s}\sim\Vec{\mathbf{S}}}}\Big(\rob(\Vec{s},0) \in \predinterval\big(\Vec{s}(0)\big)\Big) \geq 1-\alpha,
 \end{equation}
 where $\predinterval(\cdot)$ is a monitoring function producing prediction intervals guaranteed to cover the true STL robustness values of trajectories $\Vec{s}\sim\Vec{\mathbf{S}}$ with a probability greater than $1-\alpha$.
%\begin{problem}[Quantitative Predictive Monitoring]\label{prbl:qpm}
%Given a dataset $D$, 
%a significance level $\alpha\in (0,1)$ and an STL formula $\phi$, derive a monitoring function $\predinterval$ producing prediction intervals guaranteed to cover the true STL robustness values of trajectories $\Vec{s}\sim\Vec{\mathbf{S}}$ with probability greater than $1-\alpha$. Mathematically,
%\begin{equation}\label{eq:marg_cov}    \mathbb{P}_{{\Vec{s}\sim\Vec{\mathbf{S}}}}\Big(\rob(\Vec{s},0) \in \predinterval\big(\Vec{s}(0)\big)\Big) \geq 1-\alpha.
%\end{equation}
%\end{problem}
%Problem~\ref{prbl:qpm} 
The QPM problem has been solved in~\cite{cairoli2023conformal} as a conditional quantile regression problem. This boils down to learning, for a generic state $\Vec{s}(0)$, an upper and a lower quantile of the random variable $\rob(\Vec{s},0)$ induced by $\Vec{s}$. These two quantiles build the output of the function $\predinterval$ in $\Vec{s}(0)$. Conformal prediction is then used to ensure that such an interval is well-calibrated --- meaning that the probabilistic guarantees are satisfied theoretically and empirically.

A big limitation of the presented QPM problem %Problem~\ref{prbl:qpm} 
is that the monitoring function only quantifies how much the requirement is satisfied in the near future, no information about the dynamics that cause the predicted violations can be retrieved from such a monitor. Additionally, the intervals obtained in \cite{cairoli2023conformal} are not mode-specific and may hence be conservative.  We also note that the quantile regressor has to be retrained when the specification changes.} 

In this paper we tackle the aforementioned issues, and the first step to do so is to learn a surrogate model of the system dynamics --- meaning that we learn a conditional parametric distribution $p_\theta\big(\Vec{s}|\Vec{s}(0)\big)$ over observations $D_t$ that approximates the unknown distribution $p\big(\Vec{s}|\Vec{s}(0)\big)$. The surrogate model is used to gain insights into system dynamics in cases where a model of the stochastic system is either unavailable or computationally too expensive to integrate with a monitor that needs to operate efficiently in real-time.
% \iffalse
% \begin{problem}[Dynamic-aware Quantitative Predictive Monitoring]\label{prbl:dyn-qpm}
% Given a pool of observations $D_t$ from an unknown discrete-time stochastic process $\Vec{\mathbf{S}}$ over a state space $S$ with  temporal horizon $H$, 
% a significance level $\alpha\in [0,1]$ and an STL formula $\phi$, derive a trajectory generating distribution $p_\theta\big(\Vec{s}|\Vec{s}(0)\big)$ that approximates the %stochastic dynamics of the 
% stochastic system and that induces a
% monitoring function $\predinterval_\theta$ mapping a pool of approximate trajectories starting at $\Vec{s}(0)$ to an interval over robustness values such that
% \begin{equation}\label{eq:marg_cov_dyn}    \mathbb{P}_{{\Vec{s}\sim\Vec{\mathbf{S}}}}\Big(\rob(\Vec{s},0) \in \predinterval_\theta\big(\Vec{s}(0)\big)\Big) \geq 1-\alpha.
% \end{equation}
% %where $\Vec{\mathbf{s}}' = [\Vec{s}'_1,\dots ,\Vec{s}'_M]$ and  $\Vec{s}'_j\sim p_\theta\big(\Vec{s}|\Vec{s}(0)\big)$ for every $j=1,\dots, M$.
% \end{problem}
% \fi

Instead of using an end-to-end approach, i.e. directly mapping the current state to an interval over the values of the STL robustness as in~\cite{cairoli2023conformal}, we \rv{solve the QPM problem} by %will solve Problem~\ref{prbl:qpm}
leveraging deep generative models to learn the approximate distribution $p_\theta(\Vec{s}|\Vec{s}(0))$ from realizations of $\Vec{\mathbf{S}}$. In particular, we use score-based diffusion models to train a generator function $\genfnc$ acting as a distribution transformer, mapping \camera{latent samples $z$} from a standard latent distribution $p(\latent|\Vec{s}(0))$ 
%--- typically $p(\latent|\Vec{s}(0)) = \mathcal{N}(\latent;0,I)$ --- 
into samples over the target space $S^H$ (more details in Section~\ref{sec:background}). To evaluate the monitoring function $\predinterval$ 
at $\Vec{s}(0)$, we compute the empirical quantiles of the robustness values computed over trajectories sampled from $p_\theta\big(\Vec{s}|\Vec{s}(0)\big)$, followed by a calibration step based on conformal prediction.

Another important consideration is that trajectories $\Vec{s}$ may belong to different dynamical modes. Assume %there are 
\rv{that the monitored agent may choose between} $G$ different dynamical modes. A mode assignment is defined as a total function that maps trajectories into a finite set of application-specific labels $\{1,\dots, G\}$. Some modes may be rare to observe, some may always be safe, whereas others may be more prone to violations. For instance, \rv{in the crossroad scenario, depicted in Fig.~\ref{fig:multimodality}, %presents 
the agent, meaning the ego vehicle, may choose between three different dynamical modes, namely going straight, a left or right turn.} The QPM guarantees of~\eqref{eq:marg_cov} are marginal, meaning that predictors $\predinterval(\cdot)$ are guaranteed to have coverage that on average is higher than $1-\alpha$. %This means that safer modes may compensate for rare and error-prone ones. 
This means that modes with accurate predictions may compensate for rare and error-prone ones.
We aim at obtaining mode-specific predictors $\predinterval^{(\mode)}(\cdot)$ such that the coverage guarantees hold for every mode, i.e., we may require that the prediction intervals are $1-\alpha$ accurate regardless of the mode that they belong to. %Moreover, we 
We can also discern safer modes from risky ones.
To achieve such mode-conditional predictors, we assume that the calibration trajectories are labelled %with ground-truth information about 
with the dynamical mode they belong to. %This results in $G$ mode-conditional calibration sets, $D_c^{(1)}, \dots, D_c^{(G)}$. 
Mode-conditional guarantees can be stated as follows.
% \footnote{In general, if the dynamics of the stochastic process are not Markovian, one could condition on a sequence of past states, $\vec{s}(-H_p),\vec{s}(-{H_p-1}),\dots, \vec{s}(0)$, rather then on the current state $\vec{s}(0)$ alone. We focus on the current state for ease of presentation.}.

\begin{problem}[Mode-conditional Quantitative Predictive Monitoring]\label{prbl:mode-qpm}
Given two sets of trajectories $D_t$ and $D_c$ from an unknown discrete-time stochastic process $\Vec{\mathbf{S}}$ with the set of modes $\{1,\dots, G\}$ that are labelled by a, potentially unknown, mode assignment function $\mathcal{M}:S^H\to \{1,\dots, G\}$. %$\ell_{\traj}\in \{1,\dots, G\}$. 
Then, for 
a significance level $\alpha\in (0,1)$, an STL formula $\phi$, and every $\mode\in\{1,\dots,G\}$, we want to derive a mode-specific monitoring function $\predinterval^{(\mode)}$ producing intervals that cover the STL robustness of trajectories belonging to $\mode$ with a probability of no less than $1-\alpha$. Mathematically, we want to achieve
\begin{equation}\label{eq:mode_cond_guar}
\mathbb{P}_{\Vec{s}\sim\Vec{\mathbf{S}}}\Big(\rob(\Vec{s},0) \in \predinterval^{(\mode)}\big(\Vec{s}(0)\big)\ \big|\ 
\mathcal{M}(\traj) = \mode
%\ell_{\traj} = \mode 
\Big)
\geq 1-\alpha,
\end{equation}
for all $\mode\in\{1,\dots,G\}$; above,  $\mathcal{M}$ maps every realization of length $H$ of the stochastic process into one of the $G$ modes.
%where $\ell_{\traj}$ is the label identifying the mode  to which $\traj$ belongs to.
\end{problem}
To solve Problem~\ref{prbl:mode-qpm}, we will use the class-conditional version of conformal inference to obtain conditional guarantees determined by the mode-conditional calibration sets %$$D_c^{(\mode)} = \{\traj\in D_c|\ell_{\traj}=\mode\},$$ 
$ D_c^{(\mode)} = \{\traj\in D_c|\mathcal{M}(\traj) = \mode\},$
for $\mode\in\{1,\dots,G\}$. %Moreover, to build the mode-specific predictor $\predinterval^{(\mode)}$, we introduce a mode predicting function $\partition: S^H\to\{1,\dots,G\}$ mapping every realization of length $H$ of the stochastic process into one of the $G$ modes, i.e., $\partition (\Vec{s})\in\{1,\dots,G\}$. 
%This function 
In general, the mode predictor $\mathcal{M}$ may be known or learned from data either in a supervised or unsupervised fashion, depending on the availability of ground truth labels, denoting the correct mode, for a pool of trajectories. The machine-learning techniques used to tackle the problems stated above are introduced in
the next section.

%The approximate dynamics used to tackle Problem~\ref{prbl:qpm} and the mode-conditional guarantees of Problem~\ref{prbl:mode-qpm} can be combined by learning a generative model $\gendist$ that induces a mode-specific monitoring function $\predinterval_\theta^{\mode}$.

%\begin{remark}
%In general, if the dynamics of the stochastic process are not Markovian, one could condition on a sequence of past states, $\vec{s}(-H_p),\vec{s}(-{H_p-1}),\dots, \vec{s}(0)$, rather then on the current state $\vec{s}(0)$ alone. We focus on the current state for ease of presentation.
%\end{remark}

%% file: sections/background.tex
\section{BACKGROUND}\label{sec:background}

% This section outlines the key concepts from (statistical) learning theory essential for a deeper understanding of our \ourmethod\ method described in Section~\ref{sec:methods}.

\subsection{Generative Models and Diffusion Models}\label{subsec:dgm}

Every dataset can be considered as a set of observations $\mathbf{x}$ drawn from an unknown distribution $p(\mathbf{x})$. Generative models aim at learning a model that mimics this unknown distribution as closely as possible, i.e., learn a parametric distribution $p_\theta(\mathbf{x})$ as similar as possible to $p(\mathbf{x})$. The parametric distribution $p_{\theta}$ is chosen so that one knows how to efficiently sample from it. %These samples are new, meaning not previously observed, but look as if they could have belonged to the original dataset.
%The rationale of generative modeling, in general, is to generate data from noise. 
A generative model acts as a distribution transformer, i.e., a map $\gen_\theta:Z\to X$ transforming a simple distribution $p(\latent)$ over a latent space $Z$ into a complex distribution over the target space $X$. %We do not have an analytic form for such distribution but we can efficiently sample from it. 
Given a noise sample $\mathbf{z}\sim p(\mathbf{z})$, $\texttt{gen}_\theta(\mathbf{z})$ is thus a sample in $X$. 
For our monitoring application, we are interested in generating stochastic trajectories conditioned, for instance, on the starting position of the system. To approximate such conditional distribution $p(\mathbf{x}\mid \mathbf{y})$, we use 
conditional deep generative models~\cite{tashiro2021}, which can be seen as maps $\texttt{gen}_\theta: Z\times Y\to X$, where $Y$ is the conditioning space. We can generate samples in $X$ conditioned on $\mathbf{y}\in Y$, $\texttt{gen}_\theta(\mathbf{z},\mathbf{y}) \in X$, resulting in a distribution $p_\theta(\mathbf{x}|\mathbf{y})$ approximating $p(\mathbf{x}|\mathbf{y})$. 

\begin{figure}[!t]
    \centering    \includegraphics[width=0.9\textwidth]{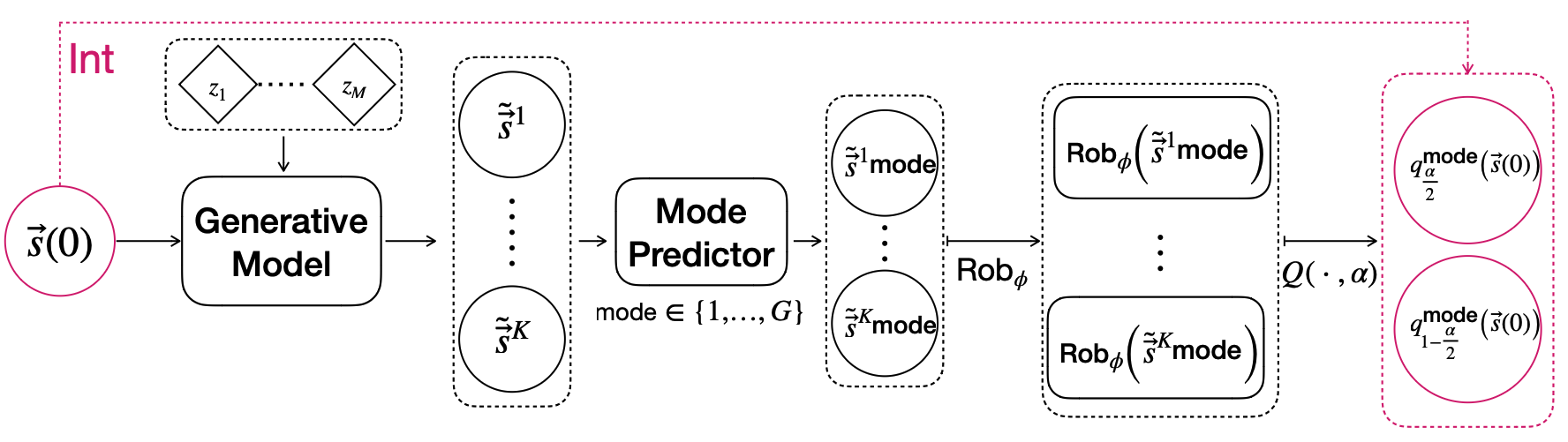}
    \vspace{-.5cm}
    \caption{ GenQPM: dynamic-aware quantitative predictive monitor with mode-conditional guarantees.}\vspace{-.4cm}
    \label{fig:mode-guar}
\end{figure}

% \paragraph{Diffusion Models.}
\rv{Diffusion models~\cite{sohldickstein2015,ho2020} are a class of generative models that learn to synthesize data by gradually denoising random noise. We focus on Denoising Diffusion Probabilistic Models (DDPMs)~\cite{ho2020}, which operate through two key processes.

\textbf{Forward Process:}
the datapoint $\mathbf{x}^0$ is sequentially corrupted by noise over a finite number of steps (indexed by $\tau$) according to a so called noise schedule $\beta_1, \dots, \beta_\mathcal{T}$:
%\begin{equation}
$
p(\mathbf{x}^\tau \mid \mathbf{x}^{\tau-1}) = \mathcal{N}\left(\sqrt{1-\beta_\tau} \mathbf{x}^{\tau-1}, \beta_\tau \mathbf{I}\right).$
%\end{equation}
After $\mathcal{T}$ steps, $\mathbf{x}^\mathcal{T}$ approximates pure noise, i.e., a Gaussian distribution $\mathcal{N}(\mathbf{0}, \mathbf{I})$.

\textbf{Reverse Process:}
a neural network $\epsilon_\theta(\mathbf{x}^\tau, \tau)$ then learns to recursively denoise $\mathbf{x}^\tau$ by predicting the noise added at each step as follows:
%\begin{equation}
$
\min_\theta\ \mathbb{E}_{\mathbf{x}^0,\epsilon,\tau}  |\epsilon - \epsilon_\theta(\mathbf{x}^\tau, \tau)|^2_2,
$%\end{equation}
where $\mathbf{x}^\tau = \sqrt{\alpha_\tau}\mathbf{x}^0 + \sqrt{1-\alpha_\tau}\epsilon$ ($\alpha_\tau := \prod_{i=1}^\tau (1-\beta_i)$). Sampling generates data by iteratively denoising $\mathbf{x}^\mathcal{T} \sim \mathcal{N}(\mathbf{0}, \mathbf{I})$.

For conditional generation (e.g., $p(\mathbf{x}^0|\mathbf{y})$), the model learns $\epsilon_\theta(\mathbf{x}^\tau, \tau \mid \mathbf{y})$ via:
$%\begin{equation}
\displaystyle\min_\theta\ \mathbb{E}_{\mathbf{x}^0,\mathbf{y},\epsilon,\tau} |\epsilon - \epsilon_\theta(\mathbf{x}^\tau, \tau \mid \mathbf{y})|^2_2.
$ %\end{equation}
A more flexible approach to controlling the generation process is \textit{guidance}~\cite{bansal2023universal,scassola2023conditioning}. }

%%-------------------

\subsection{Conformal Inference}\label{subsec:cqr}
Conformal inference~\cite{balasubramanian2014conformal,lindemann2024formal,angelopoulos2021gentle} is a lightweight statistical tool for
uncertainty quantification that can be applied on top of any supervised learning algorithm for constructing distribution-free prediction regions with guaranteed marginal coverage. The conformal framework requires minimal adaptation to suit the problem at hand. For instance, probabilistic predictors call for \emph{Conformalized Quantile Regression} (CQR)~\cite{romano2019conformalized}. Consider a probabilistic function $q(\inp)$ mapping an input $\inp\in X$ into a quantile interval over $\mathbb{R}$, i.e., $q(\inp) = [q_{lo}(\inp),\ q_{hi}(\inp)]\subseteq\mathbb{R}$. The goal is to calibrate such predictive intervals to have marginal guarantees of coverage. Starting from a dataset $D_c =\{(\inp_i,t_i)\}_i$ of pairs sampled i.i.d. from an unknown distribution $p(\inp, t)$. CQR performs the following steps:
\begin{enumerate}
    \item Define a non-conformity score function $E:X\times \mathbb{R}\to\mathbb{R}$, such that $E(\inp,t)$ quantifies the discrepancy between $t$ and the prediction interval $q(\inp)$:
    \begin{equation}\label{eq:CQR_ncs}
 E(\inp_i,t_i)=\max \{{q}_{{lo}}(\inp_i)-\targ_i,\ \targ_i-{q}_{hi}(\inp_i)\mid  (\inp_i,\targ_i)\in D_c\}.
\end{equation}
\item Use $D_c$ to define the calibration distribution 
$
\mathcal{F}_c = \sum_{i=1}^{|D_c|}\frac{1}{|D_c|+1}\delta_{e_i}+\frac{1}{|D_c|+1}\delta_{\infty},
$
where $\delta_{e}$ is the Dirac distribution with parameter $e$ and $e_i = E(\inp_i,t_i)$ is the score of the $i$-th calibration point;
\item For a given test point $\inp$ and an error rate $\alpha$, construct the prediction region as
$
C_\alpha(\inp) = \{t:E(\inp,t)\le Q(\mathcal{F}_c;1-\alpha)\},
$
where $Q(\mathcal{F}_c;1-\alpha)$ is the $1-\alpha$ quantile of $\mathcal{F}_c$. Such prediction region satisfies the following coverage guarantee w.r.t. unseen test data $(\inp,t)\sim p(\inp,t)$:
\begin{equation}
    \prob_{(\inp,t)\sim p(\inp,t)}(t\in C_\alpha(\inp))\ge 1-\alpha.
\end{equation}
\end{enumerate}
\vspace{1mm}
Note that the above holds in the more general case when $(\inp,t)$ is exchangeable w.r.t. calibration data, i.e. when the joint probability of $(\inp_1,t_1), \dots, (\inp_{|D_c|},t_{|D_c|}), (\inp, t)$ remains the same for any permutation of the data points. In practice, $C_\alpha(\inp)$ corresponds to a recalibration of the predictive interval $q(\inp)$, i.e., $C_\alpha(\inp) = [q_{lo}(\inp)-\tau_\alpha, q_{hi}(\inp)+\tau_\alpha]$, where $\tau_\alpha:=Q(\mathcal{F}_c;1-\alpha)$.\footnote{The chosen nonconformity function $E$ assumes real values and thus $\tau_\alpha$ can be negative. The conformalized prediction interval $C_\alpha(\cdot)$ can therefore be tighter than the original prediction interval $q(\cdot)$.
%This means that 
The calibrated intervals can be more efficient than the uncalibrated ones, where the efficiency is the average width of the prediction intervals over a test set.
Moreover, unlike $q(\cdot)$, $C_\alpha(\cdot)$ has guaranteed coverage.}

%\begin{remark}
%The chosen nonconformity function $E$ assumes real values and thus $\tau_\alpha$ can be negative. The conformalized prediction interval $C_\alpha(\cdot)$ can therefore be tighter than the original prediction interval $q(\cdot)$.
%This means that the calibrated intervals can be more efficient than the uncalibrated ones, where the efficiency is the average width of the prediction intervals over a test set.
%Moreover, $C_\alpha(\cdot)$ has guaranteed coverage, whereas the $q(\cdot)$ does not.
%\end{remark}
%1In the following, we will abbreviate with \textit{PI} a (non-calibrated) quantile prediction interval and with \textit{CPI} a (calibrated) conformalized prediction interval.

%% file: sections/methods.tex
\section{METHODS}\label{sec:methods}

We present \ourmethod\, a method to \rv{solve Problem~\ref{prbl:mode-qpm}.} %simultaneously solve Problems~\ref{prbl:qpm} and~\ref{prbl:mode-qpm}.
Let $\init\in\statespace$ denote the current state of the system, from which the future evolutions are stochastically distributed with probability $\condist$, which is a conditional distribution over the space of trajectories of length $\horizon$. The system's evolution should satisfy given requirements, expressed by a Signal Temporal Logic (STL) formula $\phi$. We aim to monitor such satisfaction in real-time, i.e., as the system evolves. An important consideration is that we monitor a stochastic system, meaning that satisfaction values will be stochastic too. 
% nicola: we already stated this in section 2
% We stress that the proposed methodology is not strictly related to the chosen measure of robustness. For instance, one could decide to focus on a notion of time robustness~\cite{lindemann2022temporal} and apply the same methodology described here for spatial robustness.

\iffalse
\begin{figure}[!t]
    \centering
    \includegraphics[width=\columnwidth]{imgs/genqpm_diagram.png}
    \caption{Diagram of the quantile regression scheme described in Section~\ref{subsec:alg}, i.e. the strategy to extract the two quantiles of $\distrib(\rob(\traj)\mid \init)$ given a pre-trained generative model $\gendist$, a property $\phi$ and an error level $\alpha$.}
    \label{fig:qr_diagram}\vspace{-.25cm}
\end{figure}
\fi

\input{sections/gen_qpm_alg}

\subsection{Generative QPM}\label{subsec:alg} The \ourmethod\ monitoring algorithm is outlined in Algorithm~\ref{alg:qpm} and illustrated in Fig.~\ref{fig:mode-guar}. A generative model is trained over the observed trajectories $D_t$ (step $1$) resulting in a surrogate model $\gendist$ of the system dynamics, i.e. a model that can predict the future evolutions of the system from any current state (see Section~\ref{subsec:dgm}). The surrogate generates multiple possible realizations for each calibration state (step $2.b$). For each $\mode\in\{1,\dots,G\}$, we compute the STL robustness of the surrogate trajectories mapped to $\mode$ by the mode predictor $\partition$ (step $2.c$) and use these values to extract the empirical quantiles, $q_{lo}^{\phi, \mode}(\init)$ and $q_{hi}^{\phi, \mode}(\init)$, associated respectively to $\alpha/2$ and $1-\alpha/2$ (step $2.d$). Such mode-specific quantile predictor, resulting from the composition of the generative model $\gendist$ and the mode predictor $\partition$, produces a mode-specific prediction interval $PI^{\phi, \mode}(\init) = [q_{lo}^{\phi, \mode}(\init),q_{hi}^{\phi, \mode}(\init)],$ that quantifies how much the realizations of the system belonging to $\mode$ are expected to meet the desired property $\phi$. The true robustness values of the calibration set $D_c^{(\mode,\phi)}$ (built at step $2.a$) are compared with the predictive intervals $PI^{\phi, \mode}$ to obtain a set of nonconformity scores $e_i^{\phi, \mode}$ (step $2.e$) from which we extract the $(1-\alpha)$-th empirical quantile which results in the property and mode-specific critical score $\tau^{\phi, \mode}$ (step $2.f$). This $\tau^{\phi, \mode}$ score is then used to recalibrate the predictive intervals computed over test data (step $3$) to guarantee a coverage of $1-\alpha$. The guarantees follow from the conformalized quantile regression theory introduced in Section~\ref{subsec:cqr} and state that the calibrated prediction interval $CPI^{\phi, \mode}$ is guaranteed to contain the true unknown robustness values of trajectories belonging to $\mode$
with probability $(1 - \alpha)$. It represents the monitoring function $\predinterval^{(\mode)}$ we asked for in Problem~\ref{prbl:mode-qpm}. Mathematically, 
\begin{equation}\label{eq:gen_quar}
    \prob_{\traj\sim \Vec{\mathbf{S}}}(\rob(\traj)\in CPI^{\phi, \mode}(\traj(0))\mid \mathcal{M}(\traj) = \mode)%\ell_{\traj}=\mode)
    \ge 1-\alpha,
\end{equation}
where $\traj(0)$ is the current state of the system and $\Vec{\mathbf{S}}$ denotes the random variable induced by the stochastic process $\mathbf{S} = \{\mathbf{S}(t,\omega),t\in \mathcal{I}\}$ defined before. We stress that guarantees hold under the assumption of exchangeability between calibration and test samples.

These mode-conditional guarantees allow us to obtain a range of robustness values for each possible dynamical mode. This information can be used to understand which decisions are safer than others and to identify modes that result in intervals with very high robustness, as well as those that cause potential violations. Moreover, the generative model could offer dynamics-aware insights into why the monitor is issuing an alarm. These insights can guide users in making informed decisions in response to the alarms. In particular, by analyzing the surrogate trajectories that result in negative robustness values, users can identify common undesirable behaviours potentially leading to safety violations. It is important to note that while these trajectories offer valuable insights into the causes of the alarms, they may not accurately depict potential scenarios --- the statistical guarantees hold w.r.t. the robustness values calculated from them. 
Note that the generative model $\gendist$ is trained once and for all in step (1) (independently of the property), whereas step (2) is done only once (one for each mode) after we know the property and it should be repeated only if the property changes.

\paragraph{Unbalanced generation.} The process of filtering through the mode predictor $\partition$ (step $2.c$) could lead to unbalanced datasets as the number $K_{\mode}$ of trajectories generated per mode may considerably vary based on the likelihood of observing that specific dynamics. This may affect the quality of the empirical quantiles $q_{lo}^{\phi,\mode}(\initi)$ and $q_{hi}^{\phi,\mode}(\initi)$. A solution would be to train a mode-conditional generative model so that we can generate the same number of trajectories for each mode. Mathematically, $p_\theta(\traj |\init, \mode)$ produces samples trajectores $\traj$ belonging to class $\mode\in\{1,\ldots,G\}$. However, such a solution would limit the flexibility of the generative model, as adding a dynamical mode would require a retraining of the generative model. Alternatively, one can use guidance strategies~\cite{bansal2023universal,scassola2023conditioning} to impose soft constraints over the generation of trajectories that are likely to belong to the desired mode and with no need to even retrain the model.

\begin{remark}
The mode predictor, $\partition$, can be either known a priori or learned from data. If labeled trajectories are available, where each label corresponds to the ground truth mode, supervised learning techniques can be employed to approximate the (unknown) true mode predictor. Conversely, in the absence of labels, unsupervised learning methods can be used, and the resulting learned predictor will serve as the true mode predictor.
Regardless of how $\mathcal{M}$ is obtained, the mode-specific validity condition in Equation~\eqref{eq:gen_quar} applies to the mode predictor used to partition the calibration set. However, experiments may reveal deviations from the target coverage level of $1-\alpha$ when evaluating the approximate mode predictor against a labeled test set. This occurs because the calibration data (classified by the approximate predictor) and the test data (classified by ground truth labels) lack exchangeability.
\end{remark}

The union of the $G$ mode-conditional calibrated prediction intervals, $CPI^{\phi}(\init) := \bigcup_{\mode=1}^G CPI^{\phi,\mode}(\init)$, has a guaranteed marginal coverage of $1-\alpha$ over the robustness values of all trajectories (here, using the law of total probability):
\begin{align}\label{eq:sol1_glob_cov}
   \prob_{\traj\sim \Vec{\mathbf{S}}}&\big(\rob(\traj)\in CPI^{\phi}(\init)\big) = \sum_{\mode=1}^G \prob_{\traj\sim \Vec{\mathbf{S}}}(\mathcal{M}(\traj) = \mode )%(\ell_{\traj}=\mode)
   \cdot \\
    &\cdot\prob_{\traj\sim \Vec{\mathbf{S}}}\big(\rob(\traj)\in CPI^{\phi, \mode}(\init)\mid \mathcal{M}(\traj) = \mode %\ell_{\traj}=\mode
    \big) \ge 1-\alpha.\nonumber
\end{align}

\vspace{2mm}

\iffalse
\begin{remark}
    %The mode predictor, $\partition$, can be known or learned from data, using either supervised or unsupervised methods. Notably, 
    \rv{The mode predictor, $\partition$, can be either known a priori (\textit{exact}) or learned from data (\textit{approximate}) using supervised or unsupervised methods. However, under the approximate mode predictor, the mode-specific validity in~\eqref{eq:gen_quar} applies to the predicted mode rather than the ground truth one. Consequently, inaccurate approximations may result in coverage levels that deviate from the target value of $1-\alpha$.}%remain intact regardless of the quality of $\partition$.%\fc{This remark needs better clarification.}
\end{remark}
\fi
The need for a generative model lies in that it allows approximating the conditional distribution of trajectories $\condist$, from which it is then trivial to compute \camera{empirical} upper and lower quantiles of $\distrib(\rob(\traj)\mid \init)$. An autoregressive/recurrent trajectory model, that predicts only $\mathbb{E}[\traj|\init]$ (as in~\cite{lindemann2023conformal}), does not allow us to access the full distribution and thus we cannot extract quantiles. An alternative would be to directly predict quantiles of $\distrib(\rob(\traj)\mid \init)$ (as in~\cite{cairoli2023conformal}). The advantage is that the predictor is tailored to the property and may lead to better accuracy but that is also a disadvantage in that one needs to train a predictor for any property of interest. The above approach instead requires only making a pass to the calibration distribution (step 2) whenever we know the property and need no retraining of the generative model.

\subsection{Dynamic multi-agent dynamic environments}
\begin{wrapfigure}[17]{r}{0.32\textwidth}
    \centering
    \vspace{-.7cm}
\includegraphics[width=\linewidth]{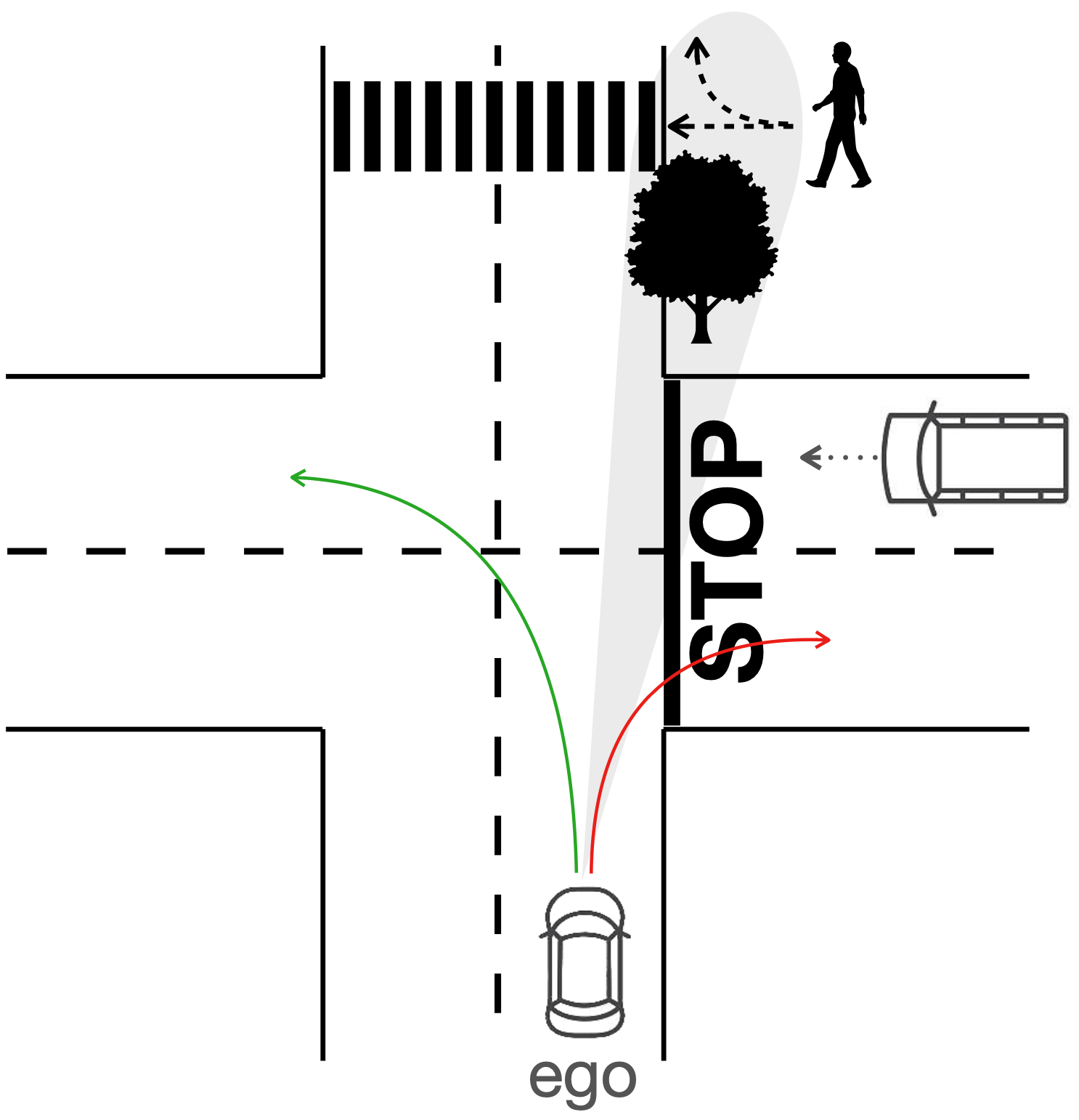}
    \vspace{-0.8cm}
    \caption{Example of an uncertain multi-agent environment: pedestrian following a stochastic behavior with an obstacle (the tree) partially hiding its position to the ego vehicle.}
    \label{fig:multimodal_crossroad}
\end{wrapfigure}

\ourmethod\ adapts efficiently to dynamically changing environments.
Consider, for instance, a scenario such as the one depicted in Fig.~\ref{fig:multimodal_crossroad}, where an ego vehicle approaches a crossroad. 
Property $\phi$ should enforce that the ego vehicle always keeps a safety distance from pedestrians and other vehicles and does not go the wrong way down a one-way street, e.g. it should not turn right from the state depicted in the figure. 
Some requirements are static, e.g., staying in the lane and not turning right. 
Some others may change dynamically over time, for instance, those when new obstacles appear or when obstacles move unpredictably. Looking at Fig.~\ref{fig:multimodal_crossroad}, we see how the car coming from the right lane could either go straight, turn left or turn right and this choice affects the safety of the monitored ego vehicle. The same applies to the pedestrian who, moreover, can become invisible to the ego vehicle when covered by the big tree.
Even when the environment changes, the mode predictor itself remains fixed (e.g., the dynamical modes of the ego vehicle are still limited to turning left, turning right, or going straight). 
However, in multi-agent environments, where interactions are highly unpredictable, the distribution of robustness values becomes significantly more complex.
%Even if the environment changes, the mode predictor does not (e.g. dynamical modes of the ego vehicle remain three turn left, right or straight). However, the distribution of robustness values gets more complex in such multi-agent unpredictable environments. 
In \ourmethod\ we face such dynamic environments as follows. \camera{For each agent that enters the scene, we train a dedicated generative model capable of emulating all possible behavioural outputs.} 
% \footnote{In our approach, each agent's surrogate is trained independently. However, including trajectories showing interactions with other agents during training, the surrogate can effectively learn and represent the reactive aspects of the system's dynamics.}. %emulatesits behaviour.
By doing so the unpredictable evolutions of the dynamic agents can be incorporated into the monitoring scheme. More precisely, as soon as the ego vehicle detects a new agent in the scene (a car, a bike or a pedestrian) it updates the property $\phi$ to account for such new agent and evaluate its robustness by simulating a pool of synthetic evolutions 
of all the agents present in the scene. As the environment dynamically evolves, we must periodically update the calibration sets $D_c^{\phi,1}, \dots, D_c^{\phi,G}$\footnote{The introduction of new agents introduces a distribution shift. While there are other ways to address distribution shifts in CP, we choose to update the calibration datasets to obtain coverage guarantees over the shifted distribution and restore exchangeability between the calibration and test set.
}  as the robustness values also depend on the unrolling of other agents' dynamics, which can be efficiently emulated to extract the mode-specific calibration scores $\tau^{\phi,\mode}$. \new{In other words, the process necessitates additional surrogate simulations for all the new agents and it involves calculating new robustness values, given that the property being monitored has changed.} 
Furthermore, once an agent is detected, \ourmethod\ continues to track it within a defined time window. This allows the monitor to account for agents temporarily obscured from view, such as a pedestrian hidden behind a tree, who might suddenly re-enter the scene and pose a collision risk. This approach enhances system safety and improves the reliability of our monitor in managing partially observable obstacles.

%% file: sections/gen_qpm_alg.tex
\begin{algorithm}[!t]
%\caption{Dynamics-Aware Quantitative Predictive Monitoring}
\begin{algorithmic}
\State \textbf{Inputs}
\begin{itemize}
    \item[-] Dataset $D_t = \{(\initi, \traj_i), i = 1,\ldots, N_t\}$ where $\initi \in S$ is the initial condition and $\traj_i$ is a trajectory of length $H$.
    \item[-] Mode-conditional calibration sets $D_c^{(\mode)}$. %$ = \{(\initi, \traj_i), i = 1,\ldots, N_c\}$.
    \item[-] Mode predictor $\partition:S^H\to\{1,\dots,G\}$
    \item[-] STL property $\phi$ to monitor.
    \item[-] Significance level $\alpha \in (0,1)$, $q_{lo}$ and $q_{hi}$ are the $\frac{\alpha}{2}$-th and $(1-\frac{\alpha}{2})$-th quantiles.
\end{itemize}

\State \textbf{Training}
\begin{enumerate}
    \item Train a generative model $\gendist$ that approximates $\condist$, the conditional distribution over the trajectory space, using $D_t$.

\item \textit{Property-specific calibration.}
 For every $\mode\in\{1,\dots,G\}$:%For every $(\initi, \traj_i) \in D_c$:
    \begin{enumerate}%[label=(\roman*)]
        
        \item Compute $D_c^{(\phi,\mode)} := \big\{(\initi, \rob(\traj_i)) \mid \traj_i \in D_c^{(\mode)}\big\}$ where $\rob(\traj_i)$ denotes the robustness of $\traj_i$ w.r.t. $\phi$.
        \item Sample $K$ trajectories $\tilde{\traj}_i^1, \ldots, \tilde{\traj}_i^K$ from $\gendisti$ for each $\initi$ from $\traj_i \in D_c^{(\mode)}$.
        \item Keep only the trajectories mapped to $\mode$ by $\partition$, i.e. keep $\tilde{\traj}_i^j$ if $\partition(\tilde{\traj}_i^j) = \mode$, and compute their robustness values $\rob(\tilde{\traj}_i^j)$ for $j \in \{1, \ldots, K_\mode\}$, where $K_\mode< K$ denotes the number of generated trajectories mapped to $\mode$ by $\partition$.
        \item Use the computed robustness values to retrieve the mode-specific lower and upper empirical quantiles:$$q_{lo}^{\phi,\mode}(\initi) = Q(\{\rob(\tilde{\traj}_i^j)\}_{j=1}^{K_\mode};\alpha/2);\  q_{hi}^{\phi,\mode}(\initi) = Q(\{\rob(\tilde{\traj}_i^j)\}_{j=1}^{K_\mode};1-\alpha/2)$$ obtaining the mode-specific prediction interval $PI^{\phi,\mode}(\initi) = [q_{lo}^\phi(\initi), q_{hi}^\phi(\initi)],$ where $Q(\cdot;\alpha)$ denotes the quantile function.
        \item Compute the calibration nonconformity scores $e_i^{\phi,\mode}$ by comparing the true $\rob(\traj_i)$ in $D_c^{(\phi,\mode)}$ with $q_{lo}^{\phi,\mode}(\initi)$ and $q_{hi}^{\phi,\mode}(\initi)$:
        $$e_i^{\phi,\mode} := \max \{q_{lo}^{\phi,\mode}(\initi)-\rob(\traj_i), \rob(\traj_i)-q_{hi}^{\phi,\mode}(\initi)\}.$$
        \item Identify $\tau^{\phi,\mode}$ as the $(1-\alpha)$-th empirical quantile of the distribution of calibration scores $\{e_i^{\phi,\mode} \mid (\initi, \rob(\traj_i)) \in D_c^{(\phi,\mode)}\}$ computed at the previous step.
    \end{enumerate}
\end{enumerate}

\State \textbf{Test}
\begin{itemize}
    \item[(3)] Given a test point $\init$, compute, for each $\mode\in\{1,\dots,G\}$, the quantile estimates $q_{lo}^{\phi,\mode}(\init)$ and $q_{hi}^{\phi,\mode}(\init)$ (as in the calibration step) and return:
$$CPI^{\phi,\mode}(\init) = [q_{lo}^\phi(\init) - \tau^{\phi,\mode},\ q_{hi}^{\phi,\mode}(\init) + \tau^{\phi,\mode}].$$
\end{itemize}
\end{algorithmic}
\caption{\textbf{\ourmethod}\ --- Illustrated in Fig.~\ref{fig:mode-guar}}\label{alg:qpm}%\vspace{-.15cm}
\end{algorithm}

%% file: sections/experiments.tex
\section{EXPERIMENTAL RESULTS}\label{sec:experiments}

We experimentally evaluate the proposed \ourmethod\ over a variety of multi-modal stochastic scenarios. We consider a one-dimensional signal showing multiple stable equilibria, a crossroad scenario, and a 2D planning scenario with obstacles to avoid. %Modes typically correspond to directions chosen by the agent. For instance, the crossroad shows three modes: continuing straight, turning left or turning right.  

\input{sections/case_studies_data}

\subsection{Case Studies}\label{subsec:case_studies}

\paragraph{\texttt{Signal}.} One-dimensional signal evolving over time showing three different stable equilibria. The STL property states that the signal will eventually reach and stay in the equilibrium with a value higher than $17.5$: $\phi = F_{[0,22]}G_{[0,22]}(\Vec{s}\ge 17.5)$. The scenario shows three dynamical modes.

\paragraph{\texttt{Navigation}.}  Agent navigating in a 2D room with square obstacles. The STL property specifies that the agent should avoid all the obstacles including the walls:
$
\phi= G_{[0,H]} 
        \big(\wedge_{i=1}^4 
        (\|\vec{\mathbf{s}}-c_i\|_\infty \ge \ell_i) \land \|\vec{\mathbf{s}}-(15,15)\|_\infty \le 14 \big)
        $
   where $(c_i, \ell_i)$ denotes the obstacles' centre and radius. The scenario shows four dynamical modes, two of which are rarely visited by the agent.

\paragraph{\texttt{Crossroad}.} Crossroad scenario shown in Fig.~\ref{fig:multimodal_crossroad}, where the ego vehicle approaches the crossroad from the bottom and there is another moving car approaching from its right.
We enforce two different STL properties: $\phi_{\texttt{right}}$ states that turning right is forbidden (as it is a one-way road),
$\phi_{\texttt{right}} = G_{[0,20]} (x \le 37),$
and $\phi_{\texttt{car}}$ states that ego should keep a safe distance from the moving car, 
$\phi_{\texttt{car}} =G_{[0,20]} \big(d(\vec{s}_{ego},\vec{s}_{car}) > 5\big). $
The scenario shows three dynamical modes. In this simpler example, we assume no pedestrian populating the scene and the other car is assumed to go straight because the no-turn signal was used to indicate a possible turn.

\paragraph{\texttt{Multi-Agent Crossroad}.} The crossroad scenario in Fig.~\ref{fig:multimodal_crossroad} features multiple obstacles with varying degrees of uncertainty. For example, the no-right-turn sign represents a static obstacle (property $\phi_{\texttt{right}}$), while the other car follows a deterministic path (property $\phi_{\texttt{car}}$). In contrast, the pedestrian’s future position is probabilistic, as they may decide to cross the street or not. Additionally, the pedestrian’s movements are only partially observable as a tree temporarily blocks the ego vehicle’s view.
The STL property is
$\phi_{\texttt{multi}} =G_{[0,20]} \Big((d(\vec{s}_{ego},\vec{s}_{ped}) > 5)\Big)\land\phi_{\texttt{car}}\land\phi_{\texttt{right}}, $
where $\phi_{\texttt{right}}$ and $\phi_{\texttt{car}}$ are the properties defined above stating respectively not to turn right and to keep a safety distance from the moving car. The dynamics of the moving obstacles are replaced by deep generative models acting as surrogates informing the ego vehicle about the possible evolutions of the multi-agent system. In this particular framework, we train a deep generative model capturing the possible choices of a pedestrian as it approaches a crossroad. 
\camera{Once the ego vehicle detects a pedestrian, it continues to track the pedestrian even if it temporarily loses sight of it for a short period.}
%From the moment the ego vehicle spots a pedestrian, it keeps it in the loop even if it is not able to observe it for a certain time window.

\begin{wrapfigure}[8]{r}{0.58\textwidth}
    \centering
       \vspace{-.8cm}    
    \includegraphics[width=0.32\linewidth]{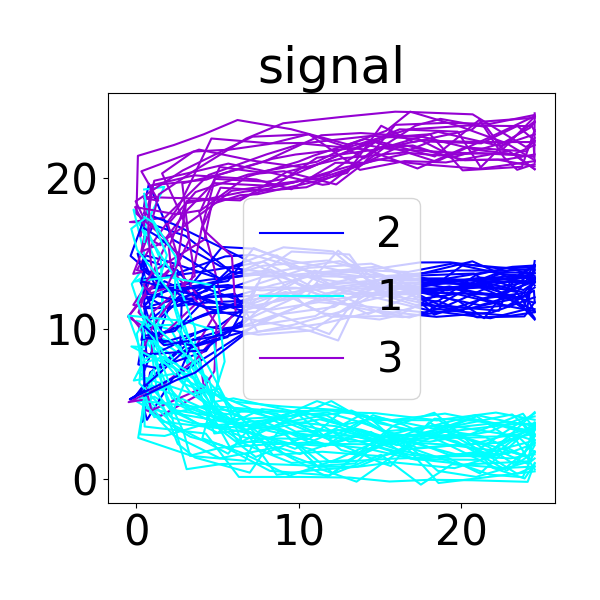}
    \hspace{-0.15cm}
\includegraphics[width=0.32\linewidth]{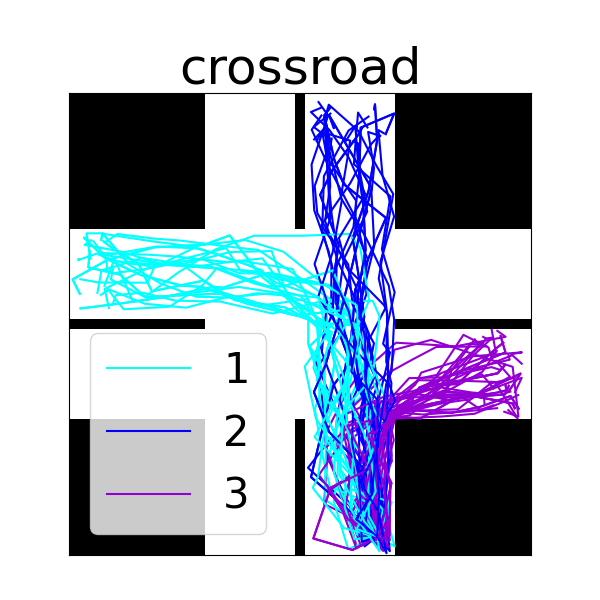}
    \hspace{-0.15cm}
\includegraphics[width=0.32\linewidth]{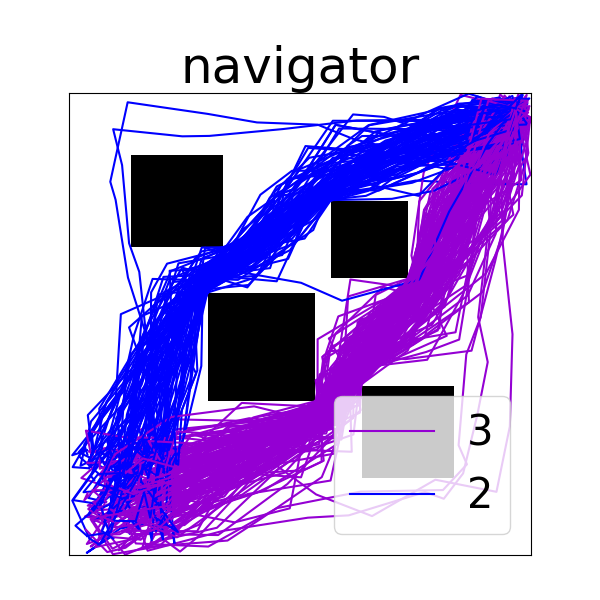}
    \vspace{-0.5cm}
    \caption{Mode predictor $\partition$ applied to trajectories from \texttt{Signal}, \texttt{Crossroad} and \texttt{Navigator} scenarios.\vspace{-.5cm}}
    \label{fig:casestudies_part}
\end{wrapfigure}
Fig.%~\ref{fig:casestudies_rob1},
~\ref{fig:casestudies_rob2} and~\ref{fig:casestudies_part} %(in Appendix~\ref{app:additional_plots}) 
show a visualization of the three case studies. Fig.%~\ref{fig:casestudies_rob1} and
~\ref{fig:casestudies_rob2} shows trajectories with the respective STL robustness values, where blue denotes higher values and red denotes lower values. 
Fig.~\ref{fig:casestudies_part} shows the same trajectories classified according to their dynamical mode. 
The \texttt{Signal} case study presents a scenario where trajectories in the same mode present similar robustness values, whereas the \texttt{Crossroad} and \texttt{Navigation} case studies present more heterogenous behaviours, where the same mode is associated with different levels of robustness.
The \texttt{Navigation} case study presents two modes that are visited less frequently than the others.
We use MATLAB Probabilistic Roadmap\footnote{\scriptsize \url{https://www.mathworks.com/help/robotics/ug/probabilistic-roadmaps-prm.html}} (PRM) ~\cite{kavraki1996probabilistic} library to generate the synthetic data of each case study.

\subsection{Evaluation Metrics}\label{subsec:eval_metrics}

We want our method to be capable of working at runtime in safety-critical applications,
which translates into the need for high reliability and high computational efficiency in producing the predictive intervals and calibrating them. We emphasize that the time required to train the generative model
does not affect the runtime efficiency, as it is performed in advance (offline) only once. The generation of trajectories evolving from the current state of the system can be easily parallelized on GPU. Such time is constant, i.e., it does not depend on the complexity of the system it approximates but only on the architecture of the generative model, and it is in general, very low. In addition, we do not want an over-conservative predictor as an unnecessarily large interval would reduce the effectiveness of our \ourmethod. Keeping that in mind, we introduce some relevant metrics to evaluate the performances of our \ourmethod. The generated quantiles are property-specific. However, monitoring a different property requires neither retraining of the generative model nor the generation of novel trajectories. We simply evaluate the quantitative STL semantics for the new property on the already generated trajectories. If a new agent appears in the scene, we generate synthetic trajectories from its current position.
The mode-agnostic version of Algorithm~\ref{alg:qpm}, where all trajectories are assumed to belong to the same mode, acts as a baseline for our experiments. 

\paragraph{Coverage and efficiency.}
We experimentally check that the guaranteed validity of \ourmethod\ is empirically met in the test evaluation. On the one hand, we compute the empirical coverage as the percentage of test trajectories whose robustness values fall inside the calibrated prediction interval. In particular, we check, for each $\mode\in\{1,\dots, G\}$, whether the empirical coverage of $CPI^{\phi,\mode}$ meets the target $1-\alpha$. On the other hand,  efficiency represents the average width of the prediction intervals over the test set. For each test input, $CPI^\phi$ contains $G$ prediction intervals. We compute the input-specific efficiency as the width of the disjoint union of such intervals and then average it over the entire test set. In general, the larger the prediction interval the more conservative the \ourmethod\ predictions. If the prediction intervals over the robustness values are always very large we have little information about the satisfaction of property $\phi$. However, the predictive efficiency must be compared with the width of the empirical quantile range (EQR), i.e. the interval that contains $(1-\alpha)$ of the simulated robustness values. We can thus measure the conservativeness as the difference in width between the predicted efficiency and the EQR width. %We also compare the coverage and the efficiency of the predicted intervals (PIs) against those of calibrated intervals (CPIs). 
Furthermore, we compare the performances under a known exact mode predictor $\partition$ with those under an approximate data-driven mode predictor (Table~\ref{tab:cov_eff_CC}). For the experiments, the approximate data-driven mode predictor is a neural network classifier trained over the training set $D_t$ labeled with the exact mode predictor.

\newcommand{\mypara}[1]{\vspace{0.6em}{\noindent {\em #1}.}}

\subsection{Experiments} The workflow can be divided into steps: (1) collect a pool of trajectories from the stochastic process, define the properties to monitor and evaluate the STL robustness over those trajectories, (2) train the generative model, (3) generate the synthetic trajectories (simulate and compute STL robustness), (4) obtain property-specific quantile intervals, (5) acquire a mode predictor $\partition$ (either knowledge-based or data-driven) (6) compute the mode-specific calibration score $\tau^{\phi,\mode}$ (obtaining calibrated prediction intervals), (7) evaluate $CPI^\phi$ on a test set.
%\ourmethod\ works only without a model of the system, in such a scenario point (2) reduces to collect a pool of trajectories and evaluate the STL robustness over those.

\mypara{Experimental settings} The entire pipeline is implemented in Python. The
generative models and the STL semantics are described in PyTorch~\cite{paszke2019pytorch}. 
The experiments were conducted on a shared virtual machine with a 32-Core Processor, 64GB of RAM and an NVidia A100 GPU with 20GB, and 8 VCPU. 
Our implementation of \ourmethod\ is available at \url{https://github.com/francescacairoli/GenerativeQPM.git}.

\mypara{Datasets} We build training sets with 3000 trajectories, calibration sets with 600 initial states and 300 trajectories per state and test sets with 200 initial states and 300 trajectories per state. \new{States are randomly sampled from a uniform distribution.  We validate the $1-\alpha$ coverage guarantees by computing for every initial state the ratio of covered robustness values for the associated 300 trajectories and then average these ratios over the 200 initial states. For each test point, we sample 500 calibration initial states in a bootstrapping manner, allowing calibration sets to vary at each test point.
}

\mypara{Training details and offline costs}
The diffusion models were trained for 200 epochs with a batch size of 512 and a learning rate of 0.0005. The same architecture and hyper-parameters for all the case studies, leveraging the stability of score-based diffusion models. The training phase takes, on average, 3 minutes. %Solution 1 is more efficient than solution 2 as the latter requires computing the prediction set $\Gamma_\partition^\epsilon$ for every generated trajectory, so we must limit the number $M$ of generated trajectories to keep the computational times low, whereas for solution 1 we simply evaluate $\partition$ over such trajectories.

\mypara{Performance evaluation} For all the experiments we choose a significance level $\alpha = 0.1$ for every mode, so that the expected coverage is $90\%$.
The lower and upper quantiles are computed respectively at $\alpha_{lo}=\alpha/2 = 0.05$ and $\alpha_{hi} = 1-\alpha/2 = 0.95$. 

\begin{figure}[!b]
\vspace{-.5cm}
    \centering
\includegraphics[width=0.9\columnwidth]{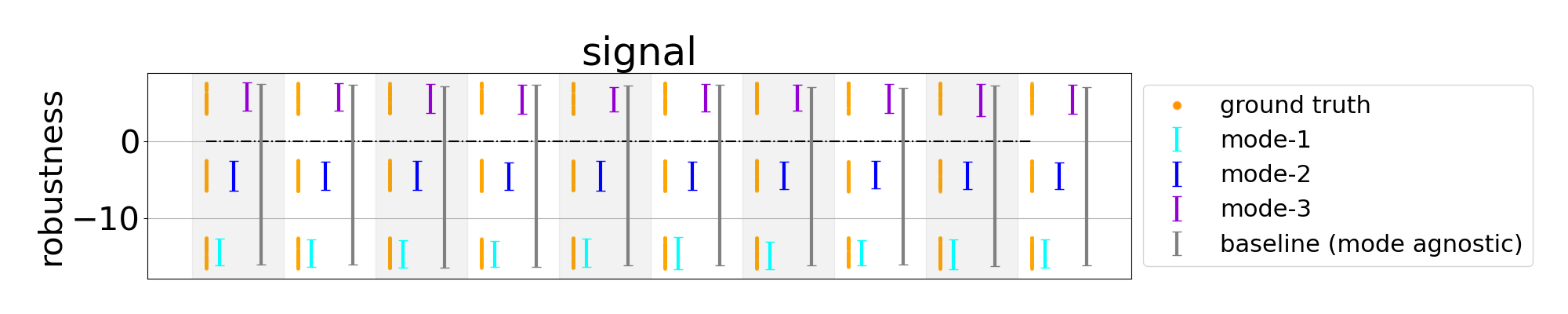}

\includegraphics[width=0.9\columnwidth]{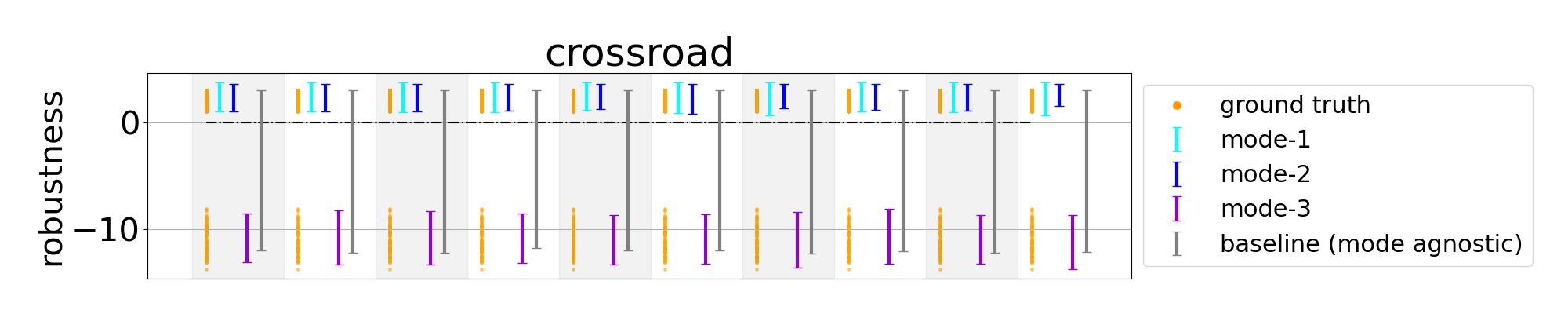}
  \vspace{-.5cm}
    \caption{ Mode-specific calibrated prediction interval with mode-specific guarantees for \texttt{Signal} (top) and \texttt{Crossroad}-$\phi_{\texttt{right}}$(bottom).}\label{fig:signal_cross1_results}\vspace{-.5cm}
\end{figure}

\begin{figure}[!t]
    \centering
    \includegraphics[width=0.9\columnwidth]{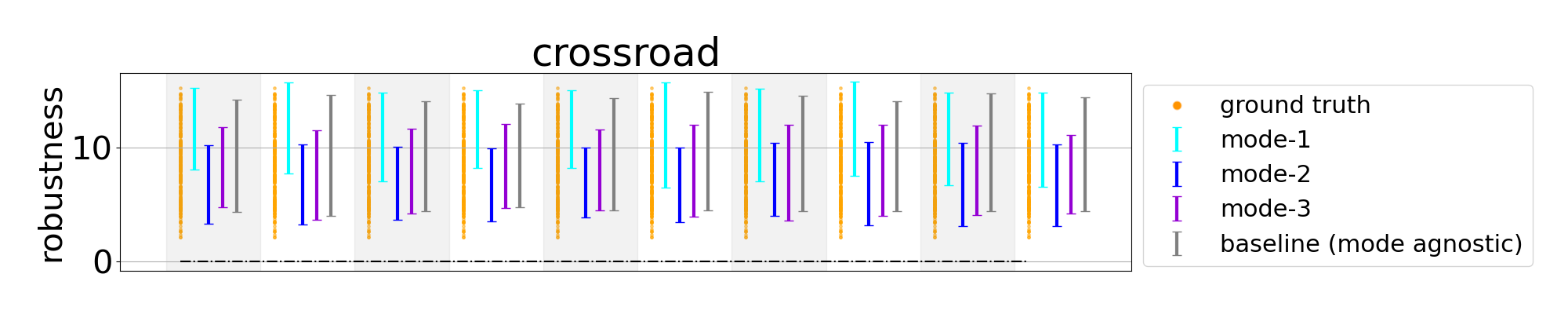}
\includegraphics[width=0.9\columnwidth]{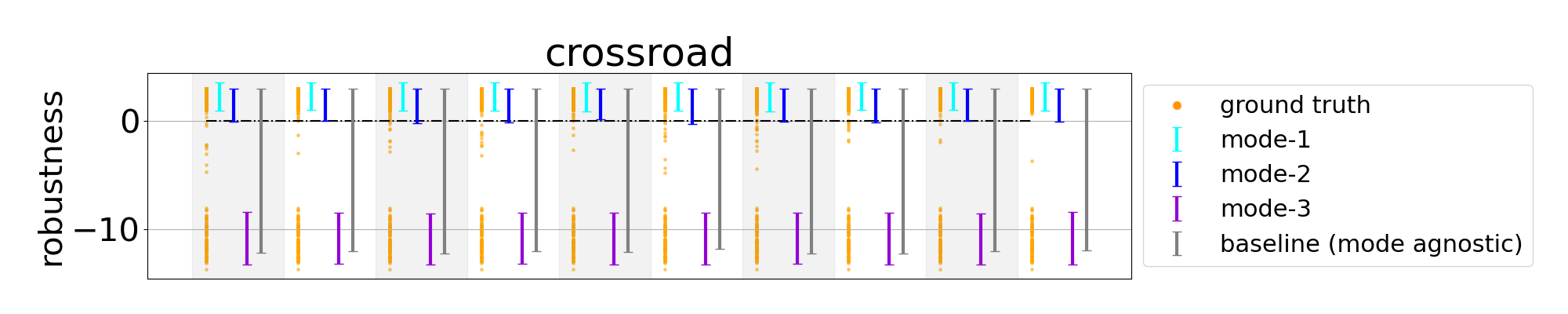}
\includegraphics[width=0.9\columnwidth]{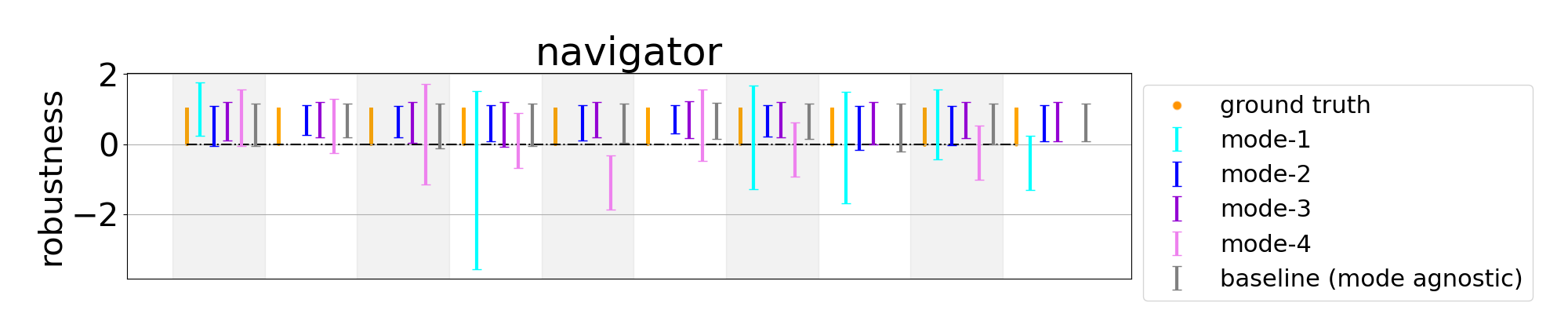}
    \vspace{-.5cm}
    %\caption{ Mode-specific calibrated prediction interval with mode-specific guarantees for  \texttt{Crossroad}-$\phi_{\texttt{car}}$(top) and \texttt{Navigator} (bottom).}\label{fig:cross2_navig_results}

%\end{figure}
%\begin{figure}[!h]
%\centering

    %\vspace{-.5cm}
    \caption{ Mode-specific calibrated prediction interval with mode-specific guarantees for \texttt{Crossroad}-$\phi_{\texttt{car}}$, $\phi_{\texttt{multi}}$ and \texttt{Navigator}.\vspace{-.5cm}}\label{fig:all_results}%\label{fig:multi-agent_crossroad_results}
\end{figure}
\subsection{Results}\label{subsec:results}
Results are summarized in 
Fig.~\ref{fig:signal_cross1_results},~\ref{fig:all_results}, %Fig.~\ref{fig:signal_results},~\ref{fig:crossroad_results},~\ref{fig:navigator_results},~\ref{fig:multi-agent_crossroad_results} 
and in Table~\ref{tab:cov_eff_CC}. %and~\ref{tab:cov_eff_approx}.
%Fig.~\ref{fig:signal_results}-~\ref{fig:multi-agent_crossroad_results} 
Fig.~\ref{fig:signal_cross1_results},~\ref{fig:all_results} compare the empirical distribution of robustness values (in orange) at 10 different initial states with the predicted CPIs. Initial states vary along the horizontal axis whereas the vertical axis represents the support of the distribution of robustness values. Each CPI comprises $G$ intervals denoted with a mode-specific colour (same colours shown in Fig.~\ref{fig:casestudies_part}).

Fig.~\ref{fig:signal_cross1_results} (top) shows the results for the \texttt{Signal} case study. The three modes of the robustness distribution coincide with the dynamical modes of the system. The three modes are well-separated and equally represented. Results show that each mode-specific CPI covers the respective peak of robustness values with the desired $90\%$ coverage. Previous approaches to QPM, as the mode-agnostic version (grey baseline interval), %(shown at the bottom of Fig.~\ref{fig:signal_results}) 
result in a unique interval spanning from $-15$ to $5$, providing no information about which is the safe mode. 

Fig.~\ref{fig:signal_cross1_results} (bottom) and Fig.~\ref{fig:all_results} (top) shows the results for the \texttt{Crossroad} case study. Property $\phi_{\texttt{right}}$ is highly bi-modal with two safe modes (going straight and turning left) and an unsafe one (the right turns of the ego vehicle). \ourmethod\ always spots the unsafe mode correctly. Property $\phi_{\texttt{car}}$ is always satisfied in the test set. Despite always being positive, mode 1 shows the highest robustness values, and it would thus be a safer choice compared to modes 2 and 3. In decision making, one could potentially combine the results of \ourmethod\  w.r.t. different STL properties and choose the mode that works better on most of the properties.

Fig.~\ref{fig:all_results} (bottom) shows the results for the \texttt{Navigator} case study. Here, mode 1 and mode 4 are under-represented in the dataset as those paths are rarely observed. \ourmethod\ balances the lower representation with higher uncertainty and thus wider prediction intervals.

Fig.~\ref{fig:all_results} (middle) %in Appendix~\ref{app:additional_plots} 
shows the results for the \texttt{Multi-Agent Crossroad} case study. The monitored requirement is a combination of $\phi_{\texttt{right}}$ and $\phi_{\texttt{car}}$ together with monitoring a safety distance from a stochastically behaving pedestrian that may sometimes cross the street. This stochastic behaviour causes a little shift towards potential violations of the safer mode.  
% The likelihood of observing an accident is low because the ego vehicle should decide to go straight when the pedestrian decides to cross. Fig.~\ref{fig:casestudies_rob1} (histogram in column 3 against the histogram in column 1), shows how the presence of a stochastically behaving pedestrian in the scene causes a little shift towards potential violations of the safer mode. 
% Fig.~\ref{fig:multi-agent_crossroad_results} shows how his shift is well 
The shift is visible in the CPI of mode 2, the mode responsible for scenarios where the ego vehicle decides to go straight. Compared to Fig.~\ref{fig:signal_cross1_results} (bottom), this interval is indeed shifted towards negative values to account for the uncertainty represented by observing a pedestrian that approaches the crossroad.

\input{tables/sol1_class_cov}

Table~\ref{tab:cov_eff_CC} %and~\ref{tab:cov_eff_approx} 
shows how CPI always meets the mode-wise desired coverage. A comparison of the efficiencies shows how \ourmethod\ is more efficient than the mode-agnostic baseline whose values are always close to the empirical quantile range (EQR). We report the percentage width gain achieved by GenQPM compared to the mode-agnostic baseline. When the robustness distribution is multi-modal, e.g. in \texttt{Signal},  \texttt{Crossroad}-$\phi_{\texttt{right}}$ and \texttt{Multi-Agent Crossroad}, the efficiency is around half w.r.t. EQR and the baseline. 
We qualitatively show how coverage (w.r.t. ground truth modes) and efficiency of CPIs obtained with a properly trained approximate mode predictor do not significantly deviate from the exact results. However, a poorly trained mode predictor may show poor performance.
Underrepresented modes, such as those in \texttt{Navigation}, result in more conservative prediction intervals. In states where the predicted interval $CPI^{\phi,\mode}$ has infinite width, we can conclude that the generative model is not properly capturing the dynamic in $\mode$, leading to under-represented calibration sets $D_c^{\phi,\mode}$ (see Sect.~\ref{subsec:alg}). Table~\ref{tab:cov_eff_CC} \rv{(\emph{Approximate} column)} illustrates how the approximate mode predictor yields prediction intervals of infinite width for these underrepresented modes. This occurs because the classifier struggles to accurately distinguish between classes with limited representation.

%% file: sections/case_studies_data.tex
\begin{figure}
\centering
\vspace{-.8cm}
\includegraphics[width=0.24\textwidth]{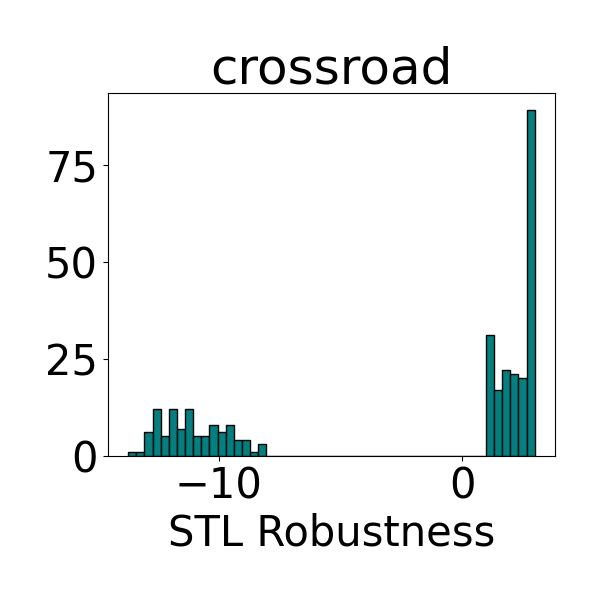}
\includegraphics[width=0.24\textwidth]{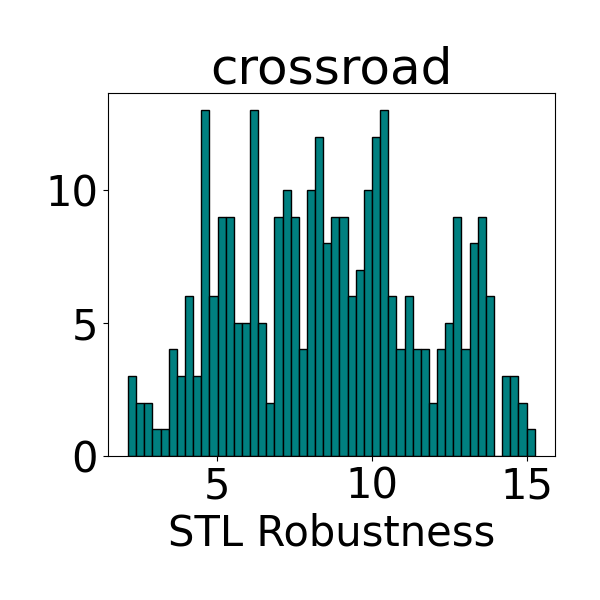}
%\includegraphics[width=0.22\textwidth]{results/crossroad_robustness_hist_prop=3.png}
%\hspace{8mm}
\includegraphics[width=0.24\textwidth]{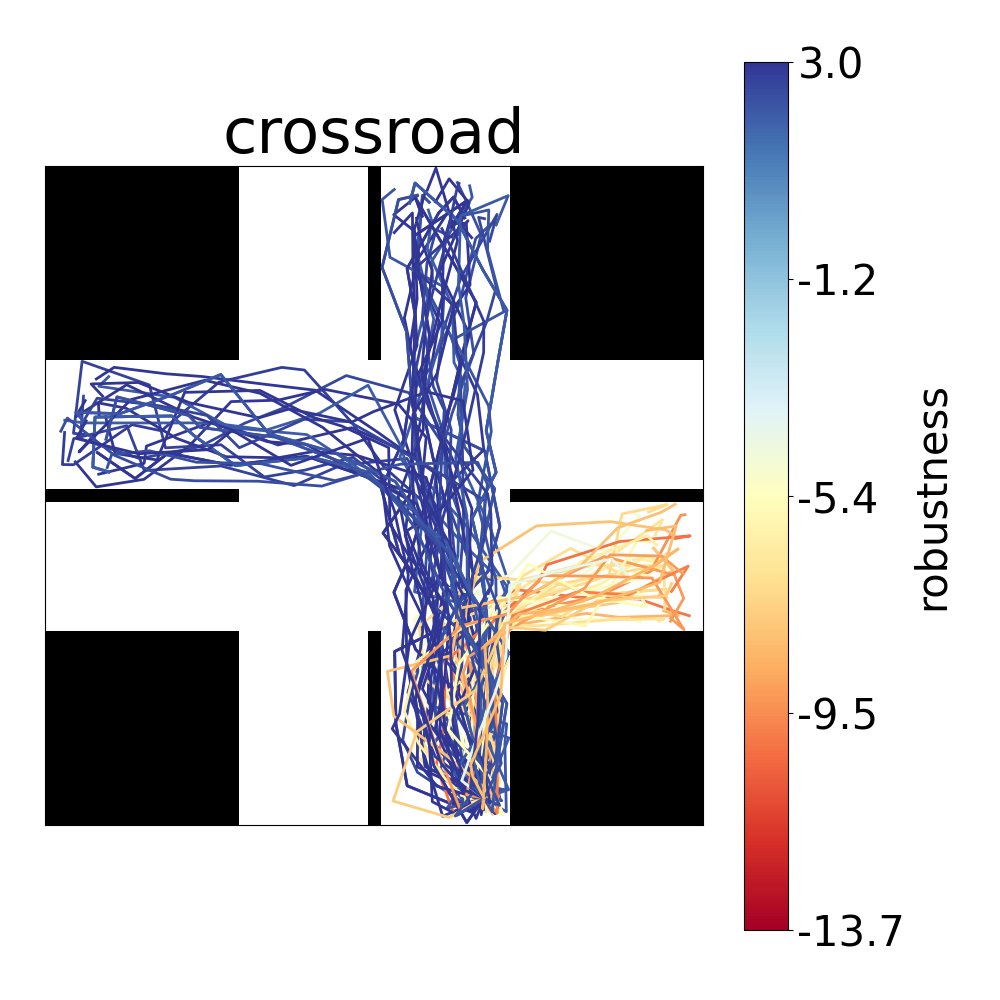}
\includegraphics[width=0.24\textwidth]{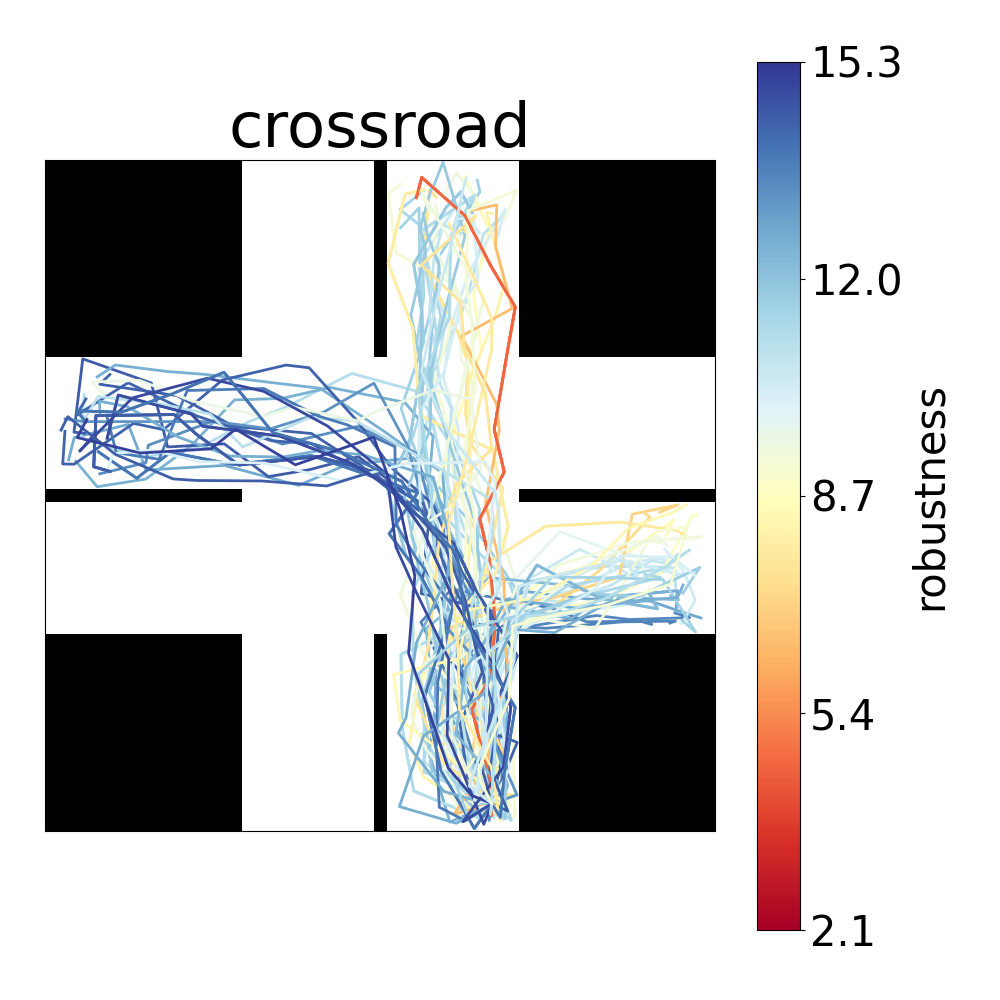}
\vspace{-.5cm}
%\caption{ Left: distribution of STL robustness values for \texttt{Crossroad}-$\phi_{\texttt{right}}$, \texttt{Crossroad}-$\phi_{\texttt{car}}$. %and \texttt{Multi-agent Crossroad} respectively. Right: Simulated trajectories with colour denoting the STL robustness (same case studies).}\label{fig:casestudies_rob1}

\includegraphics[width=0.24\textwidth]{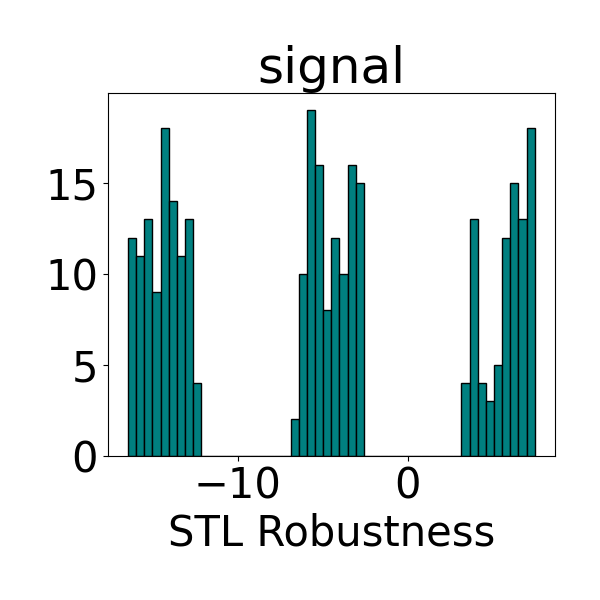}
\includegraphics[width=0.24\textwidth]{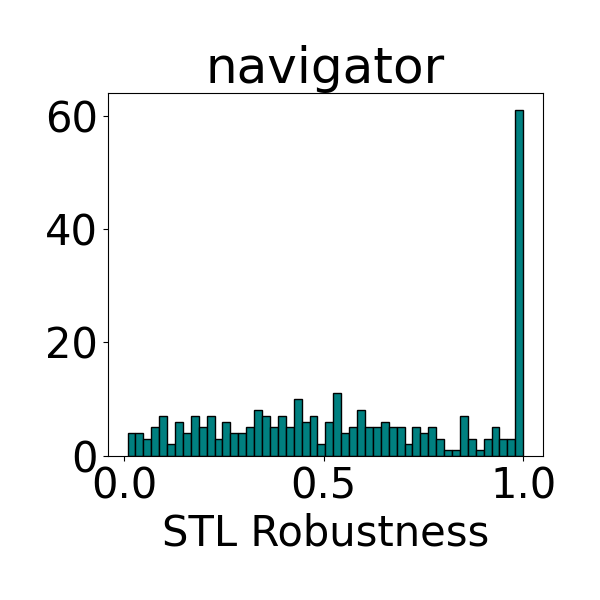}
\includegraphics[width=0.24\textwidth]{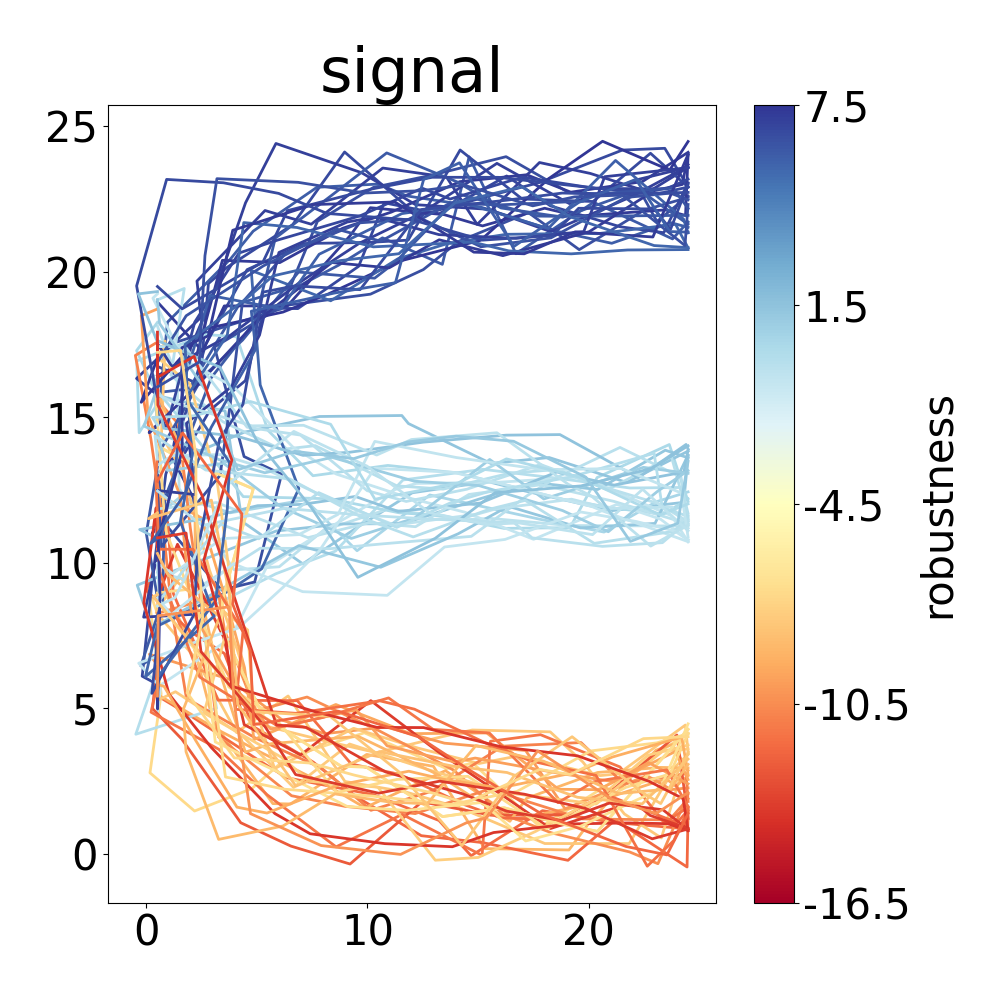}
\includegraphics[width=0.24\textwidth]{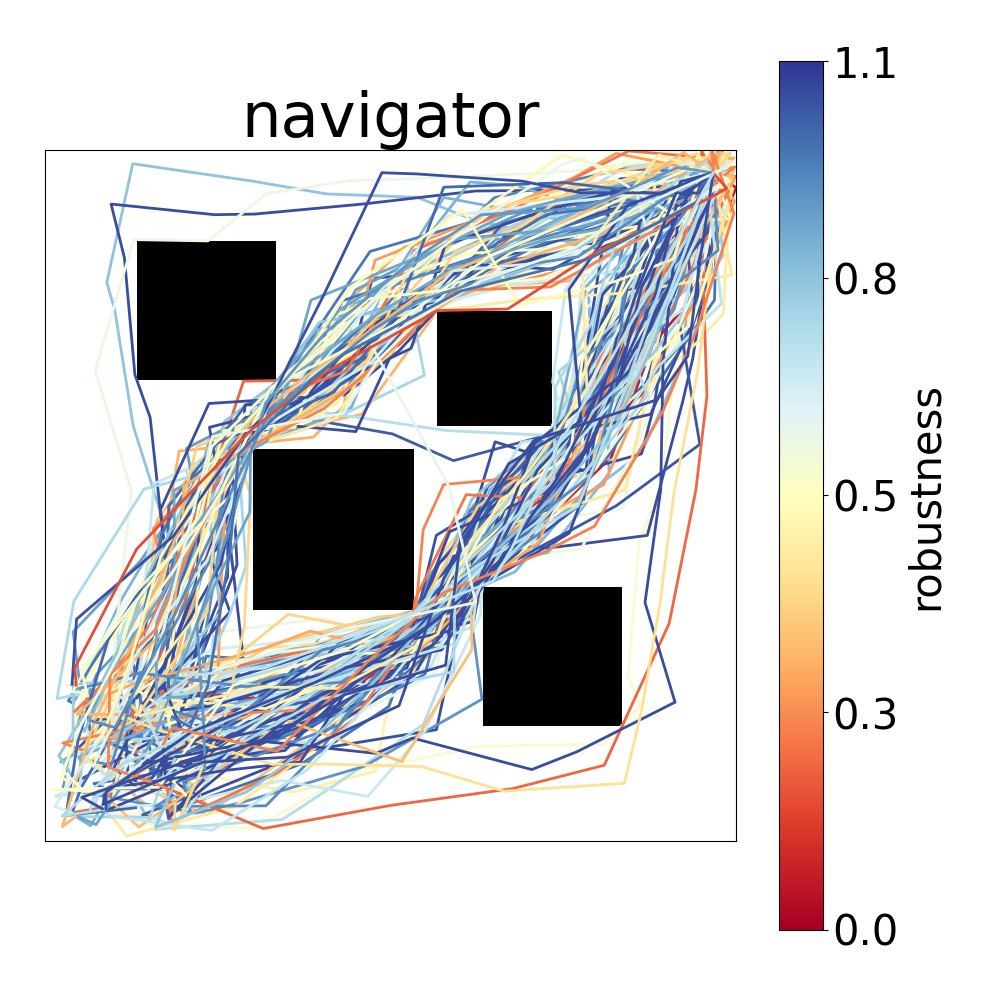}
\vspace{-.25cm}
\caption{Left: distribution of STL robustness values for \texttt{Crossroad}-$\phi_{\texttt{right}}$, $\phi_{\texttt{car}}$, $\phi_{\texttt{multi}}$, \texttt{Signal} and \texttt{Navigation} respectively. Right: Simulated trajectories with colour denoting the STL robustness for the same case studies.\vspace{-.25cm}}
\label{fig:casestudies_rob2}
\end{figure}

\iffalse
\begin{figure*}[ht]
\centering
\hspace{-10mm}
\includegraphics[width=0.19\textwidth]{results/signal_robustness_hist_prop=1.png}
\includegraphics[width=0.19\textwidth]{results/crossroad_robustness_hist_prop=1.png}
\includegraphics[width=0.19\textwidth]{results/crossroad_robustness_hist_prop=2.png}
\includegraphics[width=0.19\textwidth]{results/crossroad_robustness_hist_prop=3.png}
\includegraphics[width=0.19\textwidth]{results/navigator_robustness_hist_prop=1.png}
\includegraphics[width=0.19\textwidth]{results/signal_trajs_with_robustness_prop=1_grid.png}
\vspace{-.25cm}
\includegraphics[width=0.19\textwidth]{results/crossroad_trajs_with_robustness_prop=1_grid.png}
\includegraphics[width=0.19\textwidth]{results/crossroad_trajs_with_robustness_prop=2_grid.png}
\includegraphics[width=0.19\textwidth]{results/crossroad_trajs_with_robustness_prop=3_grid.png}
\includegraphics[width=0.19\textwidth]{results/navigator_trajs_with_robustness_prop=1_grid.png}

\caption{{\scriptsize Top: distribution of STL robustness values for \texttt{Signal}, \texttt{Crossroad}-$\phi_{\texttt{right}}$, \texttt{Crossroad}-$\phi_{\texttt{car}}$, \texttt{Multi-agent Crossroad} and \texttt{Navigation} respectively. Bottom: Simulated trajectories with colour denoting the STL robustness for the same case studies.}}
\label{fig:casestudies_rob}\vspace{-.25cm}
\end{figure*}
\fi

%% file: tables/sol1_class_cov.tex
\begin{table}[!t]
%\vspace{-.5cm}
\centering
\begin{adjustbox}{max width=\columnwidth}
\begin{tabular}{|l|r|r|r|r|r|r|r|}
\hline
\multicolumn{2}{|c|}{} & \multicolumn{2}{c|}{\textbf{Baseline}}& \multicolumn{2}{c|}{\textbf{Exact}} & \multicolumn{2}{c|}{\textbf{Approximate}}\\
\hline
\textbf{Case Study} & {EQR} & Width & Cov. & Width &  Mode-wise Coverage  & Width & Mode-wise Coverage \\
\hline
\texttt{Signal} & 22.68 & %22.75 & 90.1  
23.20 & 93.6& 10.80 {\scriptsize (-53\%)} & (90.5, 90.4, 89.9)  & 10.82 {\scriptsize (-52\%)}& (89.9, 90.2, 89.9) \\
\texttt{Xroad} - $\phi_{\texttt{right}}$ & 15.50 & 15.37 & 92.6&  7.42 {\scriptsize (-51\%)}&  (99.7, 95.3, 96.0) &  7.66 {\scriptsize (-50\%)} & (98.6, 93.6,  95.6) \\
\texttt{Xroad} - $\phi_{\texttt{car}}$ & 9.74 & %10.16 & 90.8 
10.02 & 90.0&  11.86 {\scriptsize (+18\%)}& (93.9 86.6,   89.4) & 11.92 {\scriptsize (+17\%)} &  (94.2, 87.4, 88.6)\\
\texttt{Xroad} - $\phi_\texttt{multi}$ & 15.50  & 15.03& %100
91.4 
&  8.46 {\scriptsize (-43\%)}&  (97.3, 92.6, 95.6) &  8.44 {\scriptsize (-44\%)} & (99.4,  93.3  96.0) \\
\texttt{Navigation} & 0.93 &  %1.01  & 90.3  
1.09 & 93.5&  inf & (86.2, 90.1, 90.3, 88.0) &  inf & (100., 90.7, 90.6, 100.)\\ 
\hline 
\end{tabular}
\end{adjustbox}
\caption{\textbf{Exact} and \textbf{Approximate} mode predictor vs mode-agnostic baseline: Efficiency and Coverage results across Case Studies.\vspace{-.8cm}}
\label{tab:cov_eff_CC}
\end{table}

%% file: sections/related.tex
\section{RELATED WORK}\label{sec:related}

Recent advancements in runtime predictive monitoring (PM) have seen the emergence of various learning-based approaches that employ conformal inference to provide statistical guarantees. Among these, the Neural Predictive Monitoring (NPM) framework~\cite{bortolussi2019neural,bortolussi2021neural,cairoli2021neural,cairoli2022neural,cairoli2023conformal,lindemann2023conformal} stands out for its application to a range of predictive tasks. Additionally, there has been substantial progress in reachability prediction for stochastic systems, with several methods~\cite{bortolussi2016smoothed,bortolussi2022stochastic,djeridane2006neural,royo2018classification,yel2020assured,granig2020weakness} leveraging learning techniques to address uncertainties in system behaviours. A comprehensive survey on formal verification and control algorithms for autonomous systems, which uses conformal prediction (CP) to enhance safety guarantees, can be found in~\cite{lindemann2024formal}.
Moreover, recent work has started exploring the use of CP in multimodal and dynamic environments~\cite{kiyani2024conformal,tumu2024multi,zecchin2024forking,lindemann2023safe}, expanding the applicability of these methods. Our contribution builds on these foundations by introducing a quantitative predictive monitoring approach that adapts flexibly to dynamic environments, demonstrates robust scalability, and maintains statistical guarantees. In particular, \ourmethod\ is designed to support complex stochastic systems with expressive STL-based requirements. Utilizing deep generative models, \ourmethod\ also provides interpretable insights, shedding light on potential failure causes.

%% file: sections/conclusions.tex
\section{CONCLUSIONS}\label{sec:conclusions}

We introduced \ourmethod, a learning-based technique for monitoring the behaviour of highly stochastic systems in real-time. \ourmethod\ evaluates the satisfaction of requirements—expressed as Signal Temporal Logic (STL) formulas—by computing a range of STL robustness values. These values are calibrated through principled adjustments, based on conformalized quantile regression, that ensure desired coverage levels, meaning each interval reliably captures STL robustness values with specified confidence. This feature is especially valuable for safety-critical applications, where statistical guarantees are essential.
Our method demonstrates effectiveness in complex, multimodal scenarios and provides the flexibility to monitor evolving properties without retraining the underlying generative model or generating new trajectories. This adaptability reduces runtime overhead, as the property-specific calibration score computation introduces minimal additional cost. \ourmethod\ incorporates the uncertainty related to moving hazards, e.g. external agents, by using generative models as surrogates for their dynamics. This approach allows us to estimate the potential impact of such agents on the system's safety, enhancing predictive accuracy in dynamic environments. Thanks to such generative surrogate models, \ourmethod\ provides interpretable insights into the causes of a potential imminent failure. Overall, our experimental results confirm that \ourmethod\ offers an efficient and reliable solution to the quantitative predictive monitoring problem.

%We presented \ourmethod\, a technique to reliably monitor the evolution of a highly stochastic system at runtime. In particular, given a requirement expressed as an STL formula, \ourmethod\ quantifies how robustly this requirement is satisfied by means of a range of STL robustness values. These mode-specific intervals undergo principled recalibrations that guarantee a desired level of coverage, i.e. each interval covers the exact STL robustness values with a given confidence. Such statistical guarantees our method provides play a crucial role when dealing with safety-critical applications. We show that \ourmethod\ performs well in highly multi-modal scenarios. Furthermore, one could monitor properties that change over time with no need to retrain the generative model and no need to generate novel trajectories. The property-specific cost lies in computing the property-specific calibration score causing a very little overhead during runtime execution. The experimental results support the claim of having an efficient and reliable QPM.